\documentclass[12pt]{article}

\usepackage{amsmath,amssymb}
\DeclareMathOperator*{\argmax}{arg\,max}

\usepackage{graphicx}% Include figure files
\usepackage{caption}
\usepackage{color}% Include colors for document elements
\usepackage{dcolumn}% Align table columns on decimal point
\usepackage{bm}% bold math
\usepackage{float}
\usepackage{fancyvrb}
\usepackage[colorlinks=true,linkcolor=black]{hyperref} % hyperref must be loaded before apacite
\usepackage{natbib}
\usepackage{xspace}
\usepackage{todonotes}
\usepackage[normalem]{ulem}
\usepackage{stmaryrd}
\usepackage{verbatim}

\hypersetup{ colorlinks = true, citecolor = black, urlcolor = blue}

\newcommand{\dataset}[1]{\textsc{#1}}
\newcommand{\Q}{\ensuremath{q}\xspace}

\newcommand{\FQset}{\ensuremath{\mathcal{F}}\xspace}
\newcommand{\FQ}{\ensuremath{f}\xspace}

\newcommand{\Aset}{\ensuremath{\mathcal{A}}\xspace}
\newcommand{\A}{\ensuremath{a}\xspace}

\newcommand{\KG}{\ensuremath{\mathcal{K}}\xspace}

\newcommand{\Eset}{\ensuremath{\mathcal{E}}\xspace}
\newcommand{\Lset}{\ensuremath{\mathcal{L}}\xspace}

\newcommand{\mathsc}[1]{{\normalfont\textsc{#1}}}

\newcommand{\af}[1]{\textcolor{blue}{#1}}
\newcommand{\afrm}[1]{{\color{blue}\sout{{#1}}}}

\newcommand{\gm}[1]{\textcolor{brown}{#1}}

\newcommand{\pt}[1]{\textcolor{teal}{#1}}

\newcommand{\nc}[1]{\textcolor{orange}{#1}}

\newcommand{\ncnote}[1]{\textcolor{orange}{[NOTE: {#1}]}}

\let\vec\mathbf

\definecolor{background-color}{gray}{0.98}

\title{Introduction to Neural Network based Approaches for Question Answering over 
Knowledge Graphs}
\author{Nilesh Chakraborty\thanks{Joint First Author; Smart Data Analytics Group, University of Bonn}, Denis Lukovnikov\footnotemark[1],\\ Gaurav Maheshwari\thanks{Joint First Author; Enterprise Information Systems, Fraunhofer IAIS}, Priyansh Trivedi\footnotemark[2],\\ Jens Lehmann\thanks{Smart Data Analytics Group, University of Bonn \& Enterprise Information Systems Department, Fraunhofer IAIS}, Asja Fischer\thanks{University of Bochum, Faculty of Mathematics}} 
\date{}
\begin{document}
\maketitle

\begin{center}
\subsubsection*{\small Article Type:}
Overview
% Opinion, Primer, Overview, Advanced Review, Focus Article, or Software Focus
%The Article Type denotes the intended level of readership for your article. An Editor may have mentioned a specific Article Type in your invitation letter; if so, please let them know if you think a different Article Type better suits your topic.

\hfill \break
\thanks

\subsubsection*{Abstract}
\begin{flushleft}
Question answering has emerged as an intuitive way of querying structured data sources, and has attracted significant advancements over the years. 
In this article, we provide an overview over these recent advancements, focusing on neural network based question answering systems over knowledge graphs. 
We introduce readers to the challenges in the tasks, current paradigms of approaches, discuss notable advancements, and outline the emerging trends in the field.
Through this article, we aim to provide newcomers to the field with a suitable entry point, and ease their process of making informed decisions while creating their own QA system.
% \af{[TODO: work over it again, once we are done.]}
% Question answering has emerged as an intuitive way of querying structured data sources and recently attracted significant advancements through neural and other data driven approaches. 
% In this article, we provide an overview over these recent advancements for question answering systems over structured data. 
% We introduce readers to the challenges in the tasks, current paradigm of approaches, discuss notable advancements, and outline the emerging trends in the field.
% Through this article, we aim to provide newcomers to the field with a suitable entry point, and ease their process of making informed decisions while creating their own QA system.
\end{flushleft}
\end{center}

%\clearpage 
%\tableofcontents

\clearpage

\normalsize

\clearpage

% \section*{\sffamily \Large GRAPHICAL TABLE OF CONTENTS} 
% Include an attractive full color image for the online Table of Contents. It may be a figure or panel from the article, or may be specifically designed as a visual summary. You will need to upload this as a separate file during submission.

% Size: The maximum width and height are 390 pixels, and the minimum resolution is 300 dpi. Multi-panel graphs or images are strongly discouraged.

% Caption: This is a narrative sentence to convey the article's essence and wider implications to a non-specialist audience. The maximum length is 50 words, but consider using 140 characters or less to facilitate social media sharing, which can increase the discoverability of your article.

\section{Introduction} 

% \ptt{introduce neural networks and the rise of nns}
% Introduce your topic in around 2 paragraphs, around 750 words.  Please define all acronyms at their first usage except IR, UV, NMR, DNA or similar commonly understood terms.  References should use the \href{http://www.apastyle.org/learn/quick-guide-on-references.aspx}{APA style}, like this \citep{coulson1960present}.

% \begin{enumerate}
% \item  What is question answering
% \item Why question answering 
% \item How is the survey divided (list sections)
% \item What kind of reader we are aiming for (may be)
% \end{enumerate}

% What is the field
% Why is the field important
% How is the field used
% technical introduction
% 
% What is the intended audience for the article
% Other decision points for the article
% How are advances in field useful outside?
Advancements in semantic web technologies and automated information processing systems have enabled the creation of a large amount of structured information.
% With the popularization of semantic web technologies, and automated information processing systems, a large amount of structured information is made available. 
Often, these structures follow well defined formal syntax and semantics, enabling machine readability and in some cases interoperability across different sources and interfaces.
% A prime example of such structures, and the focus of the article is a special kind of database,  which stores knowledge in the form of interconnected RDF\footnote{\url{https://www.w3.org/TR/2014/REC-rdf11-concepts-20140225/}} triples, commonly represented in the form of a knowledge graph as depicted in Fig.~\ref{fig:kg}.
%     % called knowledge graphs as depicted in Fig.~\ref{fig:kg}.
A prime example of such a structure is a special kind of graph based data model referred to as \emph{knowledge graph} (KG).
Due to their expressive nature, and the corresponding ability to store and retrieve very nuanced information, accessing the knowledge stored in KGs is facilitated through the use of formal query languages with a well-defined syntax, such as SQL, SPARQL, GraphQL etc.
% \af{Due to} their expressive nature and \af{complexity} % interfacing this 
% accessing the knowledge \af{stored in KGs is not straight forward. It} is \afrm{often} facilitated through the use of formal query languages with a well-defined syntax, such as SQL, SPARQL, GraphQL etc.
However, the use of formal queries to access these knowledge graph pose difficulties for non-expert users as % However, these interfaces provide only limited 
% \af{[Note: Where did all my chanhes go? The access is not inhibited! And the first senctence is even gramatically inccorrect!]}
% access to these knowledge sources for non-expert users as
they require the user to understand the syntax of the formal query language, as well as the underlying structure of entities and their relationships.
%is specific to every KG.
% The field of question answering over knowledge graphs (KGQA) emerged as a response to these hurdles, offering an natural guided interface to this knowledge.
% The field of question answering over knowledge graphs, emerged as a response to these hurdles, 
% attempts to interpret factual questions
% use a kg to interpret and answer factual questions
% The field of question answering over structured data, 
%     geared towards/tasked towards/focused towards
%     interpreting factual questions in natural language and answering them based on a structured data/kg source, 
%     offers an intuitive interface overcoming the aforementioned hurdles.
% To overcome the aforementioned hurdles, question answering has emerged as an intutive interface to access this knowledge by letting ()()()
%
% \ptt{Fix the following paragraph}
% As outlined by~\citeauthor{hirschman2001natural}
% \citeyear{hirschman2001natural},  
As outlined by~\citeauthor{hirschman2001natural}
(\citeyear{hirschman2001natural}),
a %question answering system over KGs,
KG question answering (KGQA) system therefore %structured data 
aims to provide the users
with an interface to ask questions in natural language, using their own terminology, to which they receive a concise answer generated by querying the KG.
% The field of 
% Question answering over structured data offers an intuitive interface % \af{strategy}
% for overcoming the aforementioned hurdles.
% As defined in~\citeNP{hirschman2001natural}}, question answering over structured data facilitates users
% \afrm{(i)} asking questions in natural language (NL) 
% \afrm{(2)} (using their own terminology) \af{and} %to which they
% \afrm{(3)} receiv\af{ing} a concise answer generated by querying a \afrm{RDF knowledge graph} \af{KG}.
Such systems have been integrated in popular web search engines like Google Search\footnote{\url{http://www.google.com/}} and Bing\footnote{\url{http://www.bing.com}} as well as in conversational assistants including Google Assistant\footnote{\url{https://assistant.google.com/}}, Siri\footnote{\url{https://www.apple.com/siri/}}, and Alexa.
% which use largescale knowledge graphs such as 
% enabled by
% enabled by largescale knowledge graphs like
% % based on 
%     Google Knowledge Graph, and Microsoft Satori respectively. 
% Further, their use in domain specific scenarios has been demonstrated to work
Their applicability in domain specific scenarios has also been %shown to work
demonstrated, for example in the
 Facebook Graph Search and IBM Watson\footnote{~\url{https://www.ibm.com/watson}}. 

Due to their wide appeal,
a variety of approaches for KGQA have been proposed. %to accomplish the KGQA task.
% a huge spectrum of various KGQA systems has been proposed. %approaches 
%ranging from traditional semantic parsing and manual feature engineering based methods to neural network based approaches that employ %generative or discriminative
%\af{classifcation, ranking or translation}
%models \afrm{to answer questions} have been proposed. 
% While 
Traditional approaches~\citep[etc.]{DBLP:conf/clef/UngerFLNCCW15,berant2013semantic,reddy2014large} typically involve a composition of template extraction, feature engineering, and traditional semantic parsing based methods.
% More recently, with the rising success of deep learning models for language understanding and related NLP tasks~\citep{DBLP:series/synthesis/2017Goldberg}, 
%     increasingly many neural KGQA approaches have been proposed.
More recently, increasingly many neural network based approaches have been shown to be effective for the KGQA task as well.
% More recently, the rising success of deep learning models for language understanding and related NLP tasks~\citep{a review}, has led to approaches that employ various deep learning models such
% neural network  %representation
% %learning 
% \af{based KGQA systems gained more an more popularity. TODO....classifcation, ranking or translation....}
%Some of these approaches include
% have led to approaches that employ various deep learning models, such as 
These approaches range from simple
neural embedding based models~\citep{bordes2014question}, over attention based recurrent models~\citep{cheng2019learning}, to memory-augmented neural controller architectures~\citep{nsm, neelakantan2016learning, yang2017differentiable}.
% \afrm{Most of these approaches solve two major subtasks either explicitly or implicitly, namely (i) understanding the entities mentioned in the question and the relations between them, and
% (ii) converting the terminology and relationships of the resources found in the question to that of the KG.} 

% \afrm{Moreover, as the field matures, a set of commonly encountered challenges along the lines of these subtasks has been identified (often, along with their workarounds) like KG artifact linking, handling out of vocabulary questions, verbalizing answers, detecting implicit relations etc. }
This article provides an introduction to neural network based methods for KGQA. 
In Section~\ref{sec:bg}, we provide the necessary background related to KGQA, introducing the terminology used in the community and the major tasks solved by KGQA systems. 
In Section~\ref{sec:datasets}, we provide a brief overview of datasets over which various KGQA systems train and benchmark their performance.
Section~\ref{sec:nnkgqa} introduces readers to the current paradigms of neural network based KGQA approaches, and provides an in-depth overview of each of these paradigms including methods for training and inference, as well as challenges specific to each paradigm, and common means to overcome them.
% \ptrm{we introduce readers to the current paradigms and discuss different approaches and the challenges they solve.} 
We conclude our discussion by listing a few emerging trends and potential future directions of research in this area in Section~\ref{sec:trends}.
% {\af{[TODO: Adapt the following after finishing the article...]}
% \af{This article provides an introduction to neural network based models for KGQA.}
%to existing literature in the field of KGQA, focusing primarily on the use of deep learning based approaches.
% We will
% (i) illustrate the primary challenges of the task
% (ii) introduce readers to the current paradigms of approaches and the challenges these approaches tackle best, and
% (iii) give an overview of the limitations of current approaches. 
We assume familiarity of readers with the basic deep learning methods and terminology and refer readers to~\citet{Goodfellow-et-al-2016} and the survey article of \citet{goldberg2016primer} for %a background 
good overviews on deep learning and applications of neural network models in natural language processing, respectively.

% ------Alternative end---
% % \pt{ A paragraph introducing the research in the field. what is the general approach. what are the challenges. maybe what paradigms of approaches}

% In this article, we aim to provide \afrm{one such} \af{an} introduction to existing literature in the field of KGQA, focusing primarily on the use of novel deep learning based approaches.
% We aim to
% (i) illustrate the \nc{primary} challenge\nc{s} \ncrm{in} \nc{of} the task
% (ii) introduce readers \ncrm{with} \nc{to} the current paradigms of approaches and the challenges these approaches tackle best, \af{and}
% (iii) give a sense of \af{the} limitations of current approaches 
% \af{[Note: We should also state somewhere which kind of prior knowledge (like about NNs) we assume the reader to have! E.g. the different kind of NNs with a reference to the DL book for readers that dont have the knowledge ]}
% \pt{ A paragraph talking about the structure of the article, what kind of audience are we writing this for etc}

% Question Answering as a field emerged alongwith these structured data sources, or knowledge bases, as an intuitive way of interfacing this knowledge.
% The goal then is to find relevant facts, and or perform basic logical operations over facts referenced in the question and present them to the user.

% challenges
%     - terminology is different
%     - kg is large what are talking about
%     - nl is not formal and to get intent and implicit links b/w things mentioned in question is difficult.

\section{Background}
\label{sec:bg}

%In this section, we discuss some 
This section  provides a brief introduction of the
concepts
necessary for an in-depth understanding of the field of KGQA, including a formal definition of KGs, a description of formal query languages, and a definition of the KGQA task and associated subtasks. 

% This section  provides  a brief introduction to KGs, formal query languages, the general KGQA task \af{and typical sub tasks.}.

\begin{figure}[h]
\includegraphics[width=\textwidth]{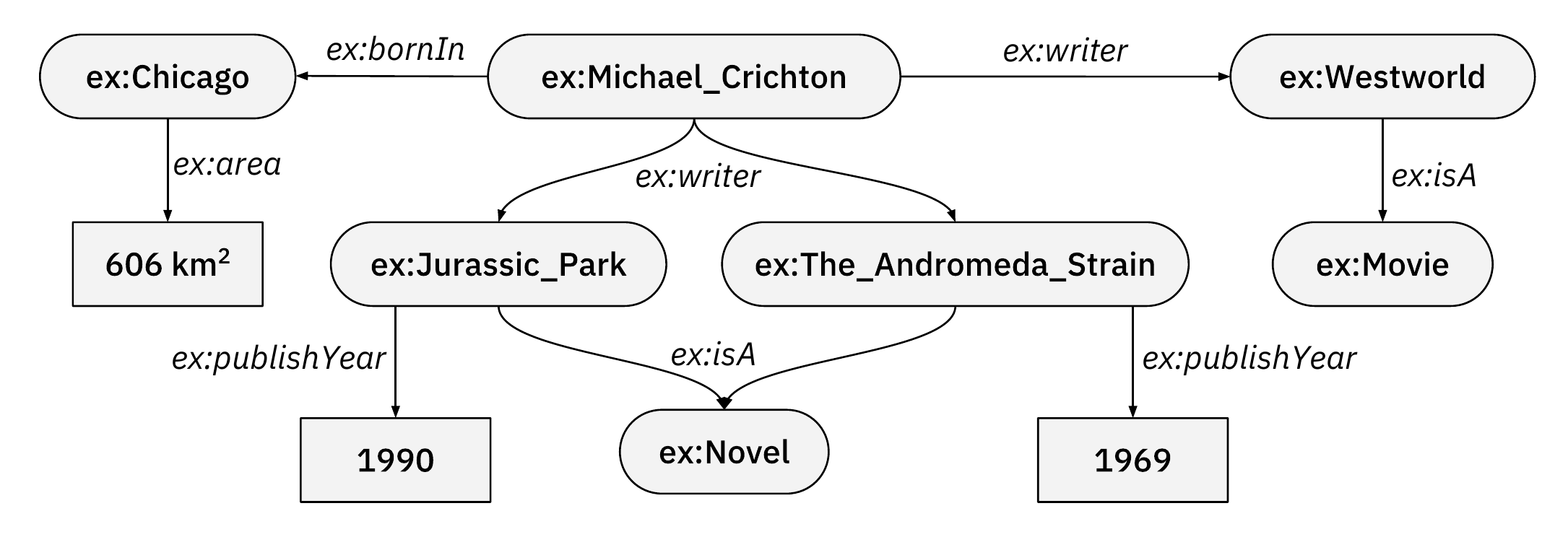}
\caption{An example KG which %we shall use 
will serve as a running example throughout this article.}
% \caption{An example of (a) knowledge graph, (b) a factual question and (c) A SPARQL query corresponding to the question, which when executed over the knowledge graph would return $\mathsf{ex{:}Chicago}$ as the result.}
\label{fig:kg}
\end{figure}

\subsection{Knowledge Graphs}
\label{sec:bg:kg}
% Knowledge graphs (KGs) form the backbone of any question answering system, and are often treated as a static source of ground truth. We begin this section by defining\ptrm{introducing} them.

% A knowledge graph is a formal representation of facts pertaining to a particular domain (including the general domain) represented as a hypergaph, whose nodes represent either,  entities denoting the subject of the interest, or literals representing atomic value of the entities like date, height etc. The edges, represented as relations, denotes the interaction between these entities.

A KG is a formal representation of facts pertaining to a particular domain (including the general domain). %collected in the form 
It consists of \textit{entities} denoting the subjects of interest in the domain, and \textit{relations}\footnote{The words \textit{predicate} or \textit{property} are often used interchangeably with relations.} denoting the interactions between these entities. 
For instance, $\mathsf{ex{:}Chicago}$ and $\mathsf{ex{:}Michael\_Crichton}$ can be entities corresponding to the real world city of Chicago, and the famous author Micheal Crichton respectively.
Further, $\mathsf{ex{:}bornIn}$ can be a relation between the two denoting that Micheal Crichton was born in Chicago. 
Instead of linking two entities, a relation can also link an entity to an data value of the entity (such as a date, a number, a string etc.)~%denoted 
which is referred to as a \textit{literal}.
The entity $\mathsf{ex{:}Chicago}$ for example
might be connected via a relation $\mathsf{ex{:}area }$ to a numerical literal with the value of $606 km^2$ describing the size of city. 
Figure~\ref{fig:kg} depicts a snippet of an example KG representing knowledge about the writer Michael Crichton, where rounded boxes denote entities, squared boxes literals, and labelled edges denote relations.

% Formally, let $\mathcal{E} = \{ e_1 \ldots e_{n_e} \}$ be the set of entities,  $\mathcal{L}$ be the set of all literal values, and $\mathcal{P} = \{p_1 \ldots p_{n_p} \}$ be the set of relations connecting two entities, or an entity with a literal.
% A knowledge graph $K$ is a subset of 
% $(\mathcal{E} \times \mathcal{P} \times (\mathcal{E} \cup \mathcal{L}))$ representing the facts that are assumed to hold.

Formally, let $\mathcal{E} = \{ e_1 \ldots e_{n_e} \}$ be the set of entities,  $\mathcal{L}$ be the set of all literal values, and $\mathcal{P} = \{p_1 \ldots p_{n_p} \}$ be the set of relations connecting two entities, or an entity with a literal.
We define a \textit{triple} $t \in \mathcal{E} \times \mathcal{P} \times (\mathcal{E} \cup \mathcal{L})$ as a statement comprising of a subject entity, a relation, and an object entity or literal, describing a fact.
For instance, one of the facts in the KG depicted in Fig.~\ref{fig:kg} can be written as
$ \langle \mathsf{ex{:}Michael\_Crichton, ex{:}bornIn, ex{:}Chicago}\rangle $. %\af{[Note: Here could be a good place to introduce labels/surface forms/names of entities and relations! Helps to understand the letter and we need to introduce it anyway]}
Then, a KG $\mathcal{K}$ is a subset of $\mathcal{E} \times \mathcal{P} \times (\mathcal{E} \cup \mathcal{L})$ of all possible triples representing facts that are assumed to hold.

\noindent
\textbf{Remark:} A common relation $\mathsf{rdfs{:}label}$ is used to link any resource (entity or relation) with a literal signifying a human readable name of the resource. For instance, $\langle \mathsf{ex{:}Michael\_Crichton,}$ $\mathsf{rdfs{:}label,}$ ``$\mathsf{Michael\  Crichton}$''@ $\mathsf{en} \rangle$. We refer to these literals as the \textit{surface form} of the corresponding resource, 
in
% throughout
the rest of this article.

% We represent the entities $\mathcal{E}$ above with a readable shorthand URI, such as $\mathsf{ex{:}Chicago}$. However, the URI are not required to be human readable. For instance, the shorthand URI for the entity corresponding to the city of Chicago, in Wikidata KG is represented as $\mathsf{wd{:}Q371938}$. 

% \ptt{Mention WIkidata, Freebase, DBpedia here and cite}

% \ptt{Add info about classes/ontology/hierarchy here? Do we need it?} We don't. If you want
% The entites and relations, belongs to classes, which itself are organized in a hiearchy defined by an ontology. are itself organized in a hiearchy 

\subsection{Formal Query Languages}
\label{sec:formalquery}
% \dl{[Note@paragraph: something is wrong. how are languages or how do languages provide mechanisms to retrieve manipulate and store data in KG??} \pt{@DL: sparql has a insert clause which can be used to populate KGs. Used it a lot in KDDS.}
A primary mechanism to retrieve and manipulate data stored in a KG is provided by %using 
formal query languages like SPARQL, $\lambda$-DCS~\citep{liang2013lambda}, or FunQL~\citep{krisp}.
These languages have a well defined formal grammar %as well as 
and structure, %and provide mechanisms 
and allow
for complex fact retrieval involving logical operands (like disjunction, conjunction, and negation), aggregation functions (like grouping or counting), filtering based on conditions, and other ranking mechanisms.
% <Add one more line to this>

\begin{figure}[h]
\includegraphics[width=\textwidth, trim={0 2.3cm 0 0},clip]{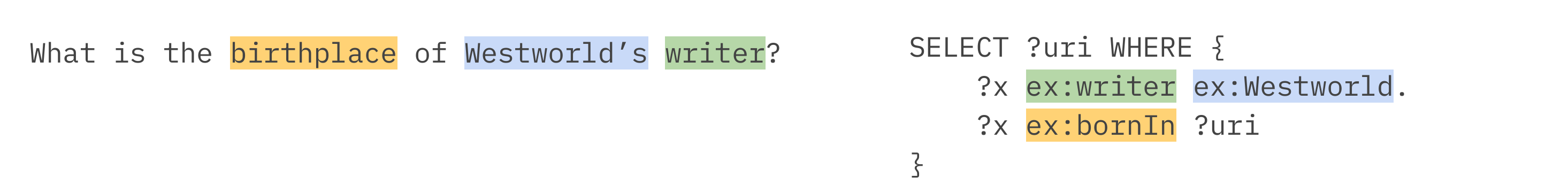}
% \caption{An example Knowledge Graph (KG) which we shall use extensively throughout the article.}
\caption{An example of (a) a (factual) NLQ and (b) a SPARQL query corresponding to the NLQ, which when executed over the KG presented in Fig.~\ref{fig:kg} would return $\mathsf{ex{:}Chicago}$ as the result.
%\af{[TODO: We should remove the sub captions in the image!]}\pt{DONE}
}
\label{fig:quesquery}
\end{figure}

SPARQL, a recursive acronym for ``SPARQL Protocol and RDF Query Language'', is one of the most commonly used query languages for KGs and is supported by many publicly available %knowledge bases
KGs like \dataset{DBpedia}~\citep{swj_dbpedia} and \dataset{Freebase}~\citep{bollacker2008freebase}. %\afrm{as well as support across several graph databases}\footnote{\url{https://en.wikipedia.org/wiki/List_of_SPARQL_implementations}}. 
For an in-depth understanding of the syntax and semantics of SPARQL, we refer interested readers to the W3C Technical Report\footnote{\url{https://www.w3.org/TR/rdf-sparql-query/\#introduction}}.
The central part of a SPARQL query is a graph, composed of resources from the KG
(i.e. entities and relations) and variables to which multiple KG resources can be mapped.
% An example of a natural language query and the corresponding SPARQL counterpart is presented in Fig.~\ref{fig:quesquery}.
As a running example that we will use throughout the rest of the paper, we consider the KG presented in Figure~\ref{fig:kg} to be the source KG over which we intend to answer the natural language query (NLQ) ``\emph{What is the birthplace of Westworld's writer?}" %by the means of the
The corresponding SPARQL query is illustrated in Figure~\ref{fig:quesquery}. 

Another popular formal query representation language is %the 
lambda calculus~\citep{introlambdacalculus}, a formal language for defining computable functions originally devised by~\citet{church1936unsolvable} and used for his study of the \textit{Entscheidungsproblem}\citep{ackermann1928grundzuge}.
The example NLQ
``What is the birthplace of Westworld's writer'' can be expressed as follows in lambda calculus:
\begin{equation}
    \lambda x . \exists y . \mathsf{bornIn}(y,x) \wedge \mathsf{writer}(x, \mathsf{Westworld}) \enspace.
\end{equation}
%
% \ncrm{
% \dl{
% \begin{equation}
%     \lambda x . \exists y . \mathsf{spouse}(x,y) \wedge \mathsf{presidentOf}(y, \mathsf{USA})
% \end{equation}
% }
% }
SPARQL and lambda calculus are both rather verbose and may result in longer and syntactically complex expressions than necessary for their use in question answering.
$\lambda$-DCS~\citep{liang2013lambda} and FunQL~\citep{krisp} both provide a more concise query representation than SPARQL or lambda calculus by avoiding variables and making quantifiers from lambda calculus implicit.
In $\lambda$-DCS, our example NLQ can be simply expressed as $\mathsf{bornIn.writer.Westworld}$.
The equivalent FunQL expression is $\mathsf{bornIn(writer(Westworld))}$.
For purposes of further exposition, we will now introduce the FunQL language in more details, adhering to the definitions provided in \citet{cheng2019learning}.
The FunQL language operates over sets. The language consists of the following components: (1) entities symbols (e.g. \textsf{Barack\_Obama}) whose denotation is the singleton set containing that entity (e.g. $\llbracket \mathsf{Barack\_Obama} \rrbracket = \{\mathsf{Barack\_Obama}\}$), (2) relation functions (e.g. \textsf{spouse()}), which return the set of entities that are reached by following the relation from its argument (e.g. $\llbracket\mathsf{spouse(Barack\_Obama)} \rrbracket = \{\mathsf{Michelle\_Obama}\}$) and (3) additional set functions, including (a) \textsf{count()}, which returns the cardinality of a set (e.g. $\mathsf{count(spouse(Barack\_Obama))} = 1$, (b) various filtering functions, which filter a set using a relation and a filtering value (e.g. $\llbracket \mathsf{filter_< (spouse(Barack\_Obama), birthdate, 1990)} \rrbracket = \{\mathsf{Michelle\_Obama}\}$), as well as (c) \textsf{argmax(), argmin()} and (d) set intersection (\textsf{and()}) and union (\textsf{or()}).

% They have a well defined grammar and syntax and provide interfaces/mechanism for complex fact retrieval involving unions and negations; complex aggregation functions like grouping or counting; filtering based on conditions, and other ranking mechanisms like ordering.
% For instance SPARQL provides a graph traversal like syntax which supports analytic query operations such as analytic query operations such as JOIN, AGGREGATE, FILTER etc. 

\subsection{%Formal definition of 
Question Answering Over Knowledge Graphs}
\label{sec:kgqa}
% [TODO: define KGQA as given question, retrieve answer and define semantic parsing as given question, retrieve logical form. $\rightarrow$ KGQA can be accomplished by semantic parsing and then execution]

% \af{[Note: Maybe we should define both KGQA taks, the standard and the weekly supervised here formally already.Then we can properöy refer to them in the approach section.]}

Using the concepts introduced above, we now define the task of KGQA.
Let \KG be a KG and let \Q be an NLQ
then we define the set of all possible answers \Aset as
%and let \Aset be the set of all possible answers that can be retrieved from \KG.
%\ncrm{and let \Aset be the set of all \nc{possible} answers a KGQA system, using \KG as ground truth, can be expected to retrieve.}
% Let $Q$ be a natural language question (NLQ) and let $\mathcal{A}$ be the set of all answers a KGQA system for knowledge graph $\mathcal{K}$ can be expected to retrieve\ncrm{, consisting of} 
the union of
(i) the power set\footnotetext{Set of all subsets of a set of elements.} $\mathcal{P}(\Eset \cup \Lset)$ of entities \Eset and literals \Lset in \KG, (ii) the set of the numerical results of all possible aggregation functions $f: \mathcal{P}(\Eset \cup \Lset) \mapsto \mathbb{R}$ (such as SUM or COUNT), and
% \ncrm{(i) a subset of entities \af{$\{e_i\}$} or literals \af{$\{e_l\}$}  from $K$, 
% (ii) the result of an arbitrary aggregation function ($f: \{e_i\} \cup \{l_i\} \mapsto \mathbb{N} $),
% or }
(iii) the set $\{\mathsf{True}, \mathsf{False}\}$ of possible boolean outcomes, which is needed for yes/no questions. % that check whether the subgraph $\Tilde{\KG}$ implied by the question \Q is contained in \KG.}
%($T/F$) variable indicating if the subgraph $\Tilde{\KG}$ implied by \Q is a subset of \KG.
The task of KGQA then is defined as returning the correct answer $a \in \mathcal{A}$, for a given question \Q.
%\dl{Please note that this is not the most general definition (for example, multi-answer questions like ``What's the capital of Ghana and is it in Africa'' are not covered) but is sufficient for most practical purposes }
%\af{[Note: 1.Isnt this what we assumed for Krantikari? But couldnt the set contain other kind of answers as well? 2. We write $a$ as an element. Should it really be a set? 3. Is it clear to the reader that all formal queries imply subgraphs? ]}

KGQA systems often
% cast as a
use a
\emph{semantic parser} to accomplish their task. Semantic parsing is the task of translating an NLQ \Q into an executable meaning representation \FQ $\in$ \FQset of \Q (see also \cite{kamath2019survey} for a recent survey on semantic parsing).
\FQset is the set of all formal queries (see Section~\ref{sec:formalquery}) that can be generated by combining entities and relations from \KG as well as arithmetic/logical aggregation functions available in the formal query language.
The correct logical form \FQ must satisfy the following conditions:
(1) the execution of \FQ on \KG 
(which we denote by $\FQ(\KG)$) % reversed pt's removal of this: we use this notation later on in weak supervision sections
yields the correct answer \A, as implied by the NLQ and 
(2) \FQ accurately captures the meaning of the question \Q.
The second condition is important because multiple logical forms could yield the expected results \A but not all of them \textit{mean} the same thing as the question \Q. %\ptrm{For example, $2+2$ and $2*2$ both execute to $4$ as the answer but only the first is the correct logical form for the question ``what is two plus two?''.} 
We call such logical forms that satisfy the first constraint but not the second \textit{spurious} logical forms.
For an example of spurious logical forms for {KGQA} lets consider the following NLQ ``What books did Michael Crichton write before 1991?''. The queries \textsf{$\mathsf{filter}_{<}$(and(isA(Novel), writtenBy(Michael\_Crichton)), publishingYear, 1991)} and \textsf{and(isA(novel), writtenBy(Michael\_Crichton))} both execute to the same set of books because Michael Crichton didn't write books after 1991. However, the second query is incorrect since it does not convey the full meaning of the question.

\paragraph{A note on terminology:}
The \textit{natural language question} (which we abbreviate to \textit{NLQ}), is also referred to as \textit{question} or \textit{utterance}. 
The \textit{meaning representation} is also referred to as \textit{logical forms} or \textit{formal queries}. 
The \textit{execution results} or \textit{answers} for a formal query are also referred to as \textit{denotations}.

\subsection{KGQA Subtasks}
\label{sec:pipeline}
Regardless of the specific semantic parsing approach taken, the construction of logical forms for KGQA requires taking several decisions regarding their structure and content.
Here, we briefly highlight those different tasks the semantic parser will have to perform.

%In this section, we introduce the primary subtasks that a KGQA system typically needs to solve in order to generate the correct formal query given a NLQ.
% \ncrm{nd challenges in the field, by means of a simple pipeline based approach, along the lines of one proposed by~\citep{dubey2016asknow,xu2014xser}.}
% \af{For an running example that} we \af{use} throughout \af{ the rest of the paper we consider the KG presented in} Figure~\ref{fig:kg} to be the source KG, \af{over} which we intend to answer the \af{NLQ} \textit{"What is the birthplace of Westworld's writer"}, by the means of the SPARQL query illustrated in Fig.~\ref{fig:quesquery}.
% TODO: problems above (i) we call it a kgqa system but we're just talking about 

% \begin{figure}[h]
% \includegraphics[width=\textwidth]{quesquery.pdf}
% % \caption{An example Knowledge Graph (KG) which we shall use extensively throughout the article.}
% \caption{An example of (a) a (factual) \af{natural language} question and (b) a SPARQL query corresponding to the question, which when executed over the \af{KG presented in} Fig.~\ref{fig:kg} would return $\mathsf{ex{:}Chicago}$ as the result.}
% \label{fig:quesquery}
% \end{figure}

% \ncrm{In order to create \af{a} formal query \afrm{(Fig.~\ref{fig:kg}})) given \af{a NLQ}, a KGQA system needs to solve the following major challenges:}
% % \af{[Note: Later on we also talk about entity span detection, so it probably be included in the list as well.]}
\begin{enumerate}
    \item \textbf{Entity Linking}: Entity linking in the context of KGQA is %finding which 
    the task of deciding which
    KG entity is referred to by (a certain phrase in) the NLQ \Q.
    In the case of our example, the entity linker must identify $\mathsf{ex{:}Westworld}$ as the entity being referred to be ``Westworld'' in the NLQ.
    The large number (often millions) of entities forms a core challenge in entity linking with large-scale KGs. 
    A particular phrase (e.g.~``Westworld''), if considered without context, can refer to several different entities (Westworld: the movie, the series, the band or the album). This phenomenon is referred to as \textit{polysemy}.
    It is essential to take the context of the phrase into account to correctly \textit{disambiguate} the phrase to the correct entity.
    %A major challenge in this task originates from the size of large-scale KGs which comprise of millions of entities.
    %Since its 
    The large number of entities makes it practically unfeasible to create fully annotated training examples to learn lexical mappings for all entities. As a consequence, statistical entity linking systems need to generalize well to unseen entities. 
    %\dl{However, learning entity lexicons as part of the model is not necessary when lexicons exist (which is the case for DBpedia, Freebase and Wikidata); possible entities for a certain phrase can then be retrieved using simple string matching. This leaves the problems of \textit{entity span detection} (identifying whether a phrase refers to any entity or none) and \textit{entity disambiguation}, both of which can be accomplished reasonably well using features that generalize well to unseen entities.}\dlnote{putting it back because entity span detection is mentioned later}
    % offer \ptrm{tremendous} \pt{significant} generalisability  \af{to not work only on a} limited scope.
    % \af{[Note:Does the letter really fit here? The approach we describe in the following does not really target generalizabilty but rather the processing of the large number of entities. If we can rather see it as a second challenge we could shift this to the end of the paragraph]}
    % A common approach  \af{of} entity linking systems \af{to handle the huge amount of KG entities,} is to \af{first} create a set of entity candidates by matching surface forms \af{[Note: did we explain before what a surface from is?]} of the entities in KG to the words in \af{the NLQ} and then exploiting the inherent structure of the KG as well as the surrounding context of the words to disambiguate/rank the candidate entities. 
Most modern KGQA systems externalize the task of entity linking by employing a standalone entity linking system like DBpedia Spotlight~\citep{mendes2011dbpedia}%, 
%EARL~\citep{dubey2018earl} % not as impactful as the other two
for DBpedia or S-Mart~\citep{yang2016s} for Freebase.
    
    \item \textbf{Identifying Relations}: 
    It is essential to determine which relation must be used in a certain part of the logical form.
    Similarly to entity linking, we need to learn to map natural language expressions to the KG - in this case to its relations. However, unlike in entity linking, where entities are expressed by entity-specific noun phrases, relations are typically expressed by noun and verb phrase \textit{patterns} that use less specific words.
    %The task of finding and linking KG relations mentioned \af{in} or implied %{implicitly referred to} 
    %\af{by} the NLQ is commonly referred to as Relation Linking.
    For instance, in our example NLQ, the relations $\mathsf{ex{:}writer}$ and $ \mathsf{ex{:}bornIn}$ are explicitly referred to by the phrases ``X's writer'' and ``the birthplace of X'', respectively.
    However, depending on the KG schema, relations may also need to be inferred, like in \textit{"Which Americans have been on the moon?"}, 
    where ``Americans'' specifies an additional constraint $\mathsf{?person\ ex{:}bornIn\ ex{:}USA}$ and $\mathsf{ex{:}bornIn}$ is never explicitly mentioned in the question.

    \item \textbf{Identifying Logical/Numerical Operators}: 
    Sometimes questions contain additional operators on intermediate variables/sets. 
    For example, "\textit{How many writers worked on Westworld}?'' implies a COUNT operation on the set of Westworld writers.
    Other operators can include ARGMAX/MIN, comparative filters (e.g. ``older than 40''), set inclusion (e.g. "\textit{Did Michael Crichton write Westworld?''}) etc.
    Like entities and relations, identifying such operators is primarily a lexicon learning problem. However, compared to entities and relations, there is a fixed small set of operators, which depends on the chosen formal language and not on the KG.
    
    %Identifying if the NLQ implies a return modifier and/or solution modifier (as described in section~\ref{sec:formalquery}), and if so, predicting the type of the modifier/s implied by the question. Typically, a question can ask for return modifiers \afrm{such as} retrieving a list of entities \afrm{(\textit{e})} or literals (like our example question), a number (e.g. "\textit{How many writers worked on Westworld}?‘‘) the validity of the fact (e.g. "\textit{Did Michael Crichton write Westworld?''}), 
    %\af{[Note: The rest of the sentence needs only consitis of fragments...]}
    %or solution modifiers like; order in the retrieve answers like highest or lowest; limit like top 10 or bottom 100 of the retrieved solutions.

    \item \textbf{Determining Logical Form Structure}: 
    %\afrm{upon linking entities, relations and fixating on a set of auxiliary constraints,} the task of constructing the formal query which upon executing on the target KG ought to fetch the desired answer.
    In order to arrive at a concrete logical form, a series of decisions must be made regarding the structure of the logical form, i.e. how to arrange the operators, relations and entities such that the resulting object executes to the intended answer.
    For example, if FunQL is used as target formal language, a decision must be taken to supply \textsf{ex:Westworld} as the argument to the relation function \textsf{ex:writer}, and to pass this subtree as the argument to \textsf{ex:bornIn}, yielding the final query \textsf{ex:bornIn(ex:writer(ex:Westworld))}.
    Note that such decisions are heavily interdependent with the previous tasks and as mentioned at the beginning of the section, the different types of decisions are not necessarily taken separately. In fact, often structural decisions are merged with or implied by lexical decisions, for example, in FunQL, generating an ARGMAX token at a certain time step implies that the next generated token will be the root of its (ARGMAX's) first argument. On the other hand, the REDUCE operation in transition-based Stack-LSTM semantic parsers~\cite{cheng2017,cheng2019learning} is a purely structure-manipulating token that indicates the finalization of a subtree in FunQL.
    
    %The primary challenge here is to fixate upon a triple pattern and conditions within the \textit{WHERE} clause of SPARQL queries.
    %For instance, given the entity and relations linked \af{to the NLQ}, they can be aggregated into valid queries in the numerous configurations.
    %\af{[Note: Can we have an example to make that clear? Furthermore, it is not clear where the following sentence is referring to!]}
    % as visualised in~\ref{fig:queryformulation}.A. 
    %A common strategy is to only consider patterns which are valid w.r.t. existing links in the KG, which when applied on the aforementioned set of patterns, suggests that pattern 1-3 are invalid.
    %In this manner, we arrive at the desired query (Fig~\ref{fig:quesquery}), which would return $\mathsf{ex{:}Michael\_Crichton}$ as the answer.
    %The \afrm{solutions} \af{approaches} to this challenge are often tightly tied together with specific KGQA solutions, the majority of which we discuss throughout the rest of this article.
    %\af{[Note: Mqybe we should not talk about solutions/approaches in this section but focus on descibing the taks/challanges?]}
    \end{enumerate}

Question answering often solve a number of these subtasks in a single process. For instance, translation based systems (see Section~\ref{sec:translation}) in principle could generate the whole query in a single sequence decoding process, thus solving all subtasks at once. However, in practice, specialized modules can be employed, for example for entity linking, in order to constrain the search space of the main model.

%A question answering system might not solve all of these subtasks in a pipeline. For instance, a seq2seq based question to logical form translation system (see Section~\ref{sec:translation}) could take the natural language question as the input, and directly return a formal query as the answer, solving all the aforementioned challenges as part of the decoding process.

% In the following section, we lay down an extensive 

\section{Datasets}
\label{sec:datasets}
Research in the field of KGQA has seen a shift from manual feature engineering based solutions~\citep{DBLP:conf/clef/UngerFLNCCW15,berant2013semantic,reddy2014large,hoffner2017survey} to neural network based, data driven approaches. One of the pivotal requirements for these data-driven approaches is the availability of large datasets comprising of a wide variety of NLQs-label pairs. As a consequence, one can observe a shift in the scale, and more recently, also in the complexity of questions in KGQA datasets.
% recently several large datasets have been released for the task.
% This shift has also reflected in the properties of datasets that have been developed over time. 
% We provide the summarised information about these datasets in Table~\ref{tab:datasets}, alongwith relevant properties pertaining to the KGQA task.
% The table~\ref{tab:datasets} describes different datasets, highlighting different properties of KGQA datasets. 
The properties of the most popular KGQA datasets are summerized in
Table~\ref{tab:datasets}.% summarises information about the \af{properties of the} most popular KGQA datasets \afrm{along with their properties relevant to the KGQA task.}
% \usepackage{booktabs}

\begin{table}[]
\resizebox{\textwidth}{!}{
\begin{tabular}{|l|l|l|l|l|}

\hline
\textbf{Dataset} & \textbf{KG} & \textbf{Size} & \textbf{Formal Queries} & \textbf{Complex Questions} \\ \hline
Free917~\citep{cai2013large}          & Freebase    & 917           & Yes                      & Yes                        \\
WebQuestions~\citep{berant2013semantic}     & Freebase    & 5810          & No                      & Yes                        \\
WebQuestionsSP~\citep{yih2016value}   & Freebase    & 4737          & Yes                     & Yes                        \\
ComplexQuestions~\citep{bao2016constraint} & Freebase    & 2100          & Yes                     & Yes                        \\
GraphQuestions~\citep{su2016generating}   & Freebase    & 5166          & Yes                     & Yes                        \\
SimpleQuestions~\citep{bordes2015large}  & Freebase    & 100K          & Yes                     & No                         \\
30M Factoid Questions~\citep{P16-1056}        & Freebase    & 30M           & Yes                     & No                         \\
QALD~\footnotemark             & DBpedia     & 50-500          & Yes                     & Yes                        \\
LC-QuAD~\citep{trivedi2017lc}           & DBpedia     & 5000          & Yes                     & Yes \\    
LC-QuAD 2~\citep{lcquad2} & DBpedia, Wikidata & 30\,000 & Yes & Yes \\
\hline                  
\end{tabular}}

\caption{(i) the underlying knowledge graph on which the dataset is based (ii) Complexity of the question (Simple/Complex) (iii) Number of questions (iv) Availability of Formal Query. 
}
\label{tab:datasets}
\end{table}
\footnotetext{https://github.com/hobbit-project/QuestionAnsweringBenchmark}

One of the first attempts to create a large scale KGQA dataset was by~\citet{cai2013large} %\citeyear{cai2013large} 
who developed  the \dataset{Free917} dataset, consisting out of 917 question / formal query pairs covering over 600 Freebase relations.
Along the same lines,~\citet{berant2013semantic} %\citeyear{berant2013semantic}
developed another dataset called \dataset{WebQuestions} by finding questions using the Google Suggest API and answering them with the help of Amazon Mechanical Turk (AMT) workers using Freebase.
Although this dataset is much larger than Free917, it suffers from two disadvantages. First, it does not provide formal queries as targets for NLQs but only % supervision on the formal queries, but instead have 
question-answer pairs, inhibiting % the use of 
supervised training of %methods  
KGQA models which rely on a logical form. Second, it is primarily composed of simple questions, and relatively few require complex reasoning.
In order to alleviate the first issue \citet{yih2016value} %&\citeyear{yih2016value} 
introduced \dataset{WebQuestionSP}, a subset of \dataset{WebQuestions}, having both answers as well as formal queries corresponding to each question.
% Although this dataset is much larger than free917, it still is largely comprised of simple question with very few (around 10 perentage)\citep{graphQuestion} ones needing complex inferencing. 
In order to provide more structural variance and expressiveness in questions,~\citet{bao2016constraint} %\citeyear{bao2016constraint}
and \citet{su2016generating} %\citeyear{su2016generating}
have released \dataset{ComplexQuestions} and \dataset{GraphQuestions}, respectively, consisting of pairs of  questions and their formal queries, where they augment a subset of \dataset{WebquestionSP} with type constraints, implicit and explicit temporal constraints, aggregation operations etc. 
\citet{trivedi2017lc}
%\citeyear{trivedi2017lc}
released \dataset{LC-QuAD}, a dataset of complex questions for the \dataset{DBpedia} KG. %~\citep{auer2007dbpedia}.
They started by generating formal queries for DBpedia and semi-automatically verbalized them using question templates, and then leveraged crowdsourcing to convert these template-based questions to NLQs,
%\ncrm{They start with formal queries, verbalising them using question templates, and then use mechanical turks to convert these template based questions to NLQs,} 
performing paraphrasing and grammar correction in the process.
% transforming it in to a natural language template and then finally using experts to convert them to natural questions.
% Apart from \dataset{LC-QuAD}, 
\citet{lcquad2} used a similar mechanism to create a significantly larger and varied dataset - \dataset{LC-QuAD 2}. 
\dataset{QALD}\footnote{QALD is an ongoing KGQA challenge with multiple tracks - see \url{http://qald.aksw.org/}.}~\citep{UsbeckGN018} is another important KGQA dataset based on DBpedia. Although it is substantially smaller than \dataset{LC-QuAD}, it has more complex and colloquial questions as they have been created directly by domain experts.
\dataset{SQA} and \dataset{CSQA} are datasets for sequential question-answering on KGs, where each data sample consists of a sequence of QA pairs with a shared context. In these datasets, while the individual questions in a sequence are generally short hilewith contextual dependencies 
% ~\citep{trivedi2017lc} created a dataset on DBpedia, \dataset{LC-QuAD}, by reverse engineering the process of Question Answering by starting with formal query, transforming it in to a natural language template and then finally using experts to convert them to natural questions. Apart from \dataset{LC-QuAD}, \dataset{QALD} is another important KGQA dataset based on DBpedia. Although it is ten times smaller than \dataset{LC-QuAD}, it has more complex and colloquial questions as they have been created directly by domain experts.
%

Most datasets discussed so far, due to their relatively small size, do not provide a thorough coverage of the entities and relations that exists in a KG.
% an end user may refer to, in asking their question }
% (Sec.~\ref{sec:approach:unseen} briefly outlines some approaches commonly taken to handle these unseen entities and relations)}{on the variety of relations in the KG}. %that could be answered using a KG. 
In order to partially compensate for this and support a larger variety of relations,~\citet{bordes2015large} %\citeyear{bordes2015large} 
created the \dataset{simple question} dataset with more that 100k questions. These questions can be answered by a single triple in the KG, and thus the corresponding formal queries can be thought to simply consist of the subject entity and predicate of the triple.
%\nc{[Note: Moved following sentence here from below. Is it relevant though? Maybe too-much-info on a single dataset. Why not report such stats for other datasets too?]\af{[Could we do this?]}}
% Interestingly, \citeauthor{petrochuk2018simplequestions} estimated an upper-bound of 83.4\% for the accuracy of KGQA on  \dataset{SimpleQuestions}, which is due to unresolvable ambiguities caused by questions lacking information to correctly disambiguate entities (i.e.~due to the existence of several triples comprising the correct relation and entity label in the KG). As current algorithms approach this limit, robust QA on simple questions appears to practically feasible given large-scale training data.   
\citet{P16-1056}
%\citeyear{P16-1056}
synthetically expanded the aforementioned dataset to create the \dataset{30M Factoid Questions} dataset consisting of 30 million question-answer pairs.
% In order to provide thorough coverage of the variety of questions that could be answered using a Knowledge base [] developed a X million simple question dataset. 
% [] synthetically expanded the dataset providing even larger coverage albeit synthetically. 
% ADD STUFF ABOUT ATIS,JOBS and THEN ABOUT WIKISQL, WikiTableQuestions and Spider

\section{%Approaches Overview
Neural Network based KGQA Systems
}
\label{sec:nnkgqa}

As explained in Section~\ref{sec:kgqa}, the problem of KGQA is usually cast into a semantic parsing problem.
 A semantic parser is an algorithm that given a context (i.e.~a KG \KG and formal target language with expressions \FQset), maps a given NLQ \Q to 
the logical form $\FQ \in \FQset$, which (1) when executed over \KG returns the correct answer \A and (2) accurately captures the meaning of \Q.
Neural network-based \textit{semantic parsing algorithms} use prediction \textit{models}, with trainable parameters, that enable the parser to be fitted on a given dataset.
The models are trained using a suitable \textit{training procedure}, which depends on the model, the parsing algorithm and the data provided for training the parser.

\begin{comment}
%\afrm{In the following sections} 
\af{In this article}\gm{,} we focus on three different types of \af{prediction models} 
%\afrm{methods} 
commonly used in semantic parsing algorithms: (i) classification, (ii) ranking\gm{,} and (iii) translation 
%\afrm{methods}
\af{models}. \af{In the following subsections (sections \ref{sec:classification}, \ref{sec:ranking}, and \ref{sec:treedecoder}, respectively) we explain each of these models,}
%\afrm{For every method, we shortly define the models},
focusing on extensions specific for semantic parsing, and discuss how they are used \af{in the parsing algorithm during inference} 
%\afrm{to perform semantic parsing} 
as well as how they are trained in the defined context.
We also provide an overview of %existing 
\af{previously proposed KGQA system which employ the different models.}%works related to th  e discussed method.
\end{comment}

In this article, we divide the prediction models commonly used in semantic parsing into three major categories, namely: (i) classification, (ii) ranking, and (iii) translation. In the following subsections (Sections~\ref{sec:classification}, \ref{sec:ranking}, and \ref{sec:translation}, respectively), we will explain each of these models and provide an overview of the previously proposed KGQA systems which employ these different models. We also discuss these models and how are they used and trained in the defined context.

In all the training procedures discussed below, the models are optimised with stochastic gradient descent (SGD) or one of its variants.
Depending on the type of model, different loss functions can be used. 
Moreover, depending on the semantic parsing algorithm, different ways of processing and using the provided data in the optimization algorithm are required in order to train the models.
These choices define the different training procedures that are discussed in the subsequent sections.
Another important aspect 
%\afrm{of training semantic parsers} 
is the type of training data which is provided, leading to two different training settings:
the \textit{fully supervised} setting, where
the dataset consists of $N$ pairs of NLQs and formal queries $%\mathcal{D} =
\left\{ \left( \Q^{(i)}, \FQ^{(i)} \right) \right\}_{i=1}^N$, and the \textit{weakly supervised} setting, where the semantic  parser is trained over a dataset of pairs of NLQs and corresponding execution results $%\mathcal{D} = 
\left\{ \left( \Q^{(i)}, a^{(i)} \right) \right\}_{i=1}^N$.
Whereas the 
%\afrm{modelling aspect of the discussed methods is}
prediction models themselves are generally not affected by this difference, it does lead to different training procedures.

%\nc{[Note: Moving weaksup intro stuff here from the translation section.]}
% \afrm{Building large-scale datasets of questions annotated with logical forms is an expensive process. 
% \dl{It is generally considered cheaper to create datasets by annotating questions with their answers (execution results) rather than a full logical form\nc{~\citep{berant2013semantic}}. }
% However, training semantic parsers using only answers (weakly supervised training) requires a change in the training procedure (w.r.t. the fully supervised training) since we can no longer maximize the likelihood function of the correct logical forms (they are not given).}
%
While the fully supervised setting allows to train the models by simply maximizing the likelihood of predicting the correct logical form, the weakly supervised setting presents a more challenging scenario, where we must indirectly infer and ``encourage'' the (latent) logical forms that execute to the correct answer while avoiding spurious logical forms which might hurt generalization. This gives rise to two fundamental challenges. Firstly, finding \textit{consistent} logical forms (those which execute to the ground truth answer) is challenging due to the size of the search space which usually grows exponentially with respect to the number of tokens in the logical form. Secondly, %\afrm{in addition to the sparsity of consistent logical forms}%\footnote{There is huge number of possible queries that can be constructed with a large knowledge graph. However, only a few of them will execute to the correct answer.}, 
%\afrm{we also need to} 
dealing with spurious candidates, that is, incorrect logical forms (those that do not capture the meaning of the source question) that coincidentally execute to the correct answer, thereby misleading the supervision signal provided to the semantic parser~\citep{cheng2018Weakly}.

\subsection{Classification based KGQA}
\label{sec:classification}

%\afrm{In semantic parsing for \af{KGQA}, the input is a \af{NLQ} in the form of a sequence of tokens \af{(e.g.~words or characters [Note: what to add?]}).}
In the most general case, the semantic parser should be able to generate a structured output (i.e.~the  corresponding formal query) of arbitrary size and complexity %based on 
given the input NLQ. %\ncrm{we would like to generate a structured object based on this input, where the objects can have different sizes and structures for different inputs.}
In some cases, however, we can assume a fixed structure for the formal query. This holds in the case of single fact based questions (e.g.~\dataset{SimpleQuestions}), where only a single subject entity and a relation need to be predicted.
For instance, the question \emph{"Where was Michael Crichton born?"} is represented in our example KG (Fig.~\ref{fig:kg}) by a single triple pattern based SPARQL query with the triple pattern being $\langle \mathsf{ex{:}Michael\_Crichton, ex{:}bornIn, ?x}\rangle $. The logical form corresponding to this type of question thus always consists of one subject entity and a relation, and the answer is given by the missing object entity.
%\footnote{\af{A task closely related to KGQA with formal queries following a fixed structure is the \dataset{WikiSQL} task, where an SQL query needs to be generated for a given NLQ. In this task, the \textsc{SELECT} clause always contains two arguments, the \textsc{SELECT} column and an aggregator, which can be predicted using separate classifiers~\citep{sqlnet} (see for example \citet{lukovnikov2018translating} and \citet{huang2018natural}).}}
%\af{[Note: Remove the following since its not really KGQA?? Or shift it to a footnote?]}%\afrm{\dl{Another example is the \dataset{WikiSQL} task, where an SQL query needs to be generated from a NLQ. In the \dataset{WikiSQL} task, the \textsc{SELECT} clause always contains two arguments, the \textsc{SELECT} column and aggregator, which can be predicted using separate classifiers~\cite{sqlnet}.}}
For such fixed-structure prediction tasks, we can make use of simple text classification methods to predict the different parts of the target formal query given the NLQ as input (see also \cite{hakimov2019evaluating} for a recent overview of deep learning architectures for simple questions).

\subsubsection{Classification Models}%\ptrm{\af{To illustrate such classification methods lets take a look at the task of relation classification given an NLQ $\Q$, i.e.~the task of predicting which of the $n_p$ relations $p_1,\dots, p_{n_p} \in \mathcal{P}$ of the underlying KG $\mathcal{K}$ $\Q$ is referring to.}}
To illustrate such classification methods, let us consider the task of relation classification. Given an NLQ $q$, and a KG $\mathcal{K}$, the relation classification task consists of predicting which of the $n_r$ relations $r_1 ,\ldots, r_{n_r} \in P_{\mathcal{K}}$ is referred to in $q$. In the first step, an \emph{encoder network} can be used to map
the variable-length input sequence of $\Q$ onto a fixed-length  vector $\vec{q} \in \mathbb{R}^d$,
which is called \emph{the latent representation}, or the \emph{encoded representation} of $\Q$. 
%\af{[Note: Keep in mind, that you removed the word "embedding" here! If we keep with that decision, take care that its nowhere used in the context of NLQs or relations etc!]}
The encoder network can be comprised of different neural architectures, including recurrent neural networks (RNN)~\citep{DBLP:journals/neco/HochreiterS97}, convolutional neural networks (CNN)~\citep{lecun1998convolutional}, or transformers~\citep{transformer}.
The encoded question is subsequently fed through an affine transformation to calculate a \emph{score vector} ${s (\Q)}=(s_1(\Q) \dots s_{n_r}(\Q))$, as follows:
\begin{equation}\label{eq:af_trans}
   {s}(\Q) = W_o \vec{q} + \vec{b_o}  \enspace.
\end{equation}
%\ptrm{\af{to calculate a \emph{score vector} ${s (\Q)}=(s_1(\Q) \dots s_{n_p}(\Q))$,}wh}
Here, $W_o\in \mathbb{R}^{n_r \times d}$ and $\vec{b_o} \in \mathbb{R}^{n_r}$, in conjunction with the parameters of the encoder network, constitute the trainable parameters of the classification model.
In the output layer the classification model typically turns the score vector into a
conditional probability distribution $ p(r_k | \Q)$ over the $n_r$ relations
based on a softmax function, that is
\begin{equation}\label{eq:softmax}
    p(r_k| \Q) = \frac{e^{s_k(\Q)}}{\sum_{j=0}^{n_r} e^{s_j(\Q)}} \enspace,
\end{equation}
for $k=1,\dots n_r$. Classification is then performed by picking the relation with the highest probability given $\Q$.

%\paragraph{Training}
Given a data set of $N$ pairs of NLQs and single fact based formal queries $\mathcal{D} =
\left\{ \left( \Q^{(i)}, \FQ^{(i)} \right) \right\}_{i=1}^N$, (for \dataset{SimpleQuestions}, $\FQ^{(i)}$ is a entity-relation tuple: $\FQ^{(i)} = (e^{(i)}, r^{(i)})$),
%[TODO: DENIS, later we say the data set consists out of triples, we need to make this consistent! ]
the relation classification model is trained by maximising the log-likelihood of the model parameters $\theta$, which is given by 
\begin{equation}
 \sum_{i=1}^N  \log  p_{\theta}(r^{(i)}| \Q^{(i)}) \enspace,
\end{equation}
where $r^{(i)}$ is the predicate used in the formal query $\FQ^{(i)}$.

\subsubsection{Classification based Parsing Algorithms} For single-fact based prediction tasks, systems %based purely 
relying
on standard classification models as described above, can achieve state-of-the-art performance~\citep{mohammed2018,petrochuk2018simplequestions}.
%\citet{mohammed2018} explore different baselines as well as neural network-based classifiers (CNNs and several RNN variants) %for relation \af{classification} 
%\af{and}
%\citet{petrochuk2018simplequestions} \af{bidirectional LSTMs}~\citep{schuster1998bidirectional} for relation classification. %and achieve state-of-the-art end results.
Since the target formal queries consist only of one subject entity and one relation, they can, in principle, be predicted using two separate classifiers, receiving the NLQ as input and producing an output distribution over all entities and all relations in the KG,
respectively.
This approach can be successfully applied to predict the KG relation mentioned or implied in the question.
However, large KGs like Freebase contain a huge amount of entities and training datasets can only cover a small fraction of these (and therefore many classes remain unseen after training). This makes it difficult to apply the approach described above for relation classification to entity prediction.\footnote{Relation prediction might suffer from the same problem as entity prediction. However, usually there are much fewer relations (thousands) than entities (millions) in a KG, such that a dataset of substantial size 
can provide several instances of each (important) relation.}
%
%this naive approach does not scale to large KGs like Freebase since KGQA training datasets  can cover only a small fraction of their huge amount of entities (and therefore many classes remain unseen after training). 
%
Therefore,
several works on \dataset{SimpleQuestions} initially only perform relation classification and rely  on a two-step approach for entity linking instead. In such an two-step approach, first an entity span detector\footnote{The entity span detector does not need to learn representations for entities (like encoder networks used in classifiers), avoiding the need for data covering all entities.} is applied to identify the entity mention in the NLQ. 
Secondly, 
%\afrm{given this entity mention, a disambiguation approach is used to select the best \af{fitting} subject entity from the KG.}
%The entity span detector does not need to learn representations for entities, avoiding the need for data covering all entities. 
%After predicting the entity span, 
simple text based retrieval methods can be used to find a limited number of suitable candidate entities.
The best fitting entity given $\Q$ can then be chosen from this candidate set based on simple string similarity and graph-based features.
%\af{[Note: From the narrative, dont we miss here the explaination why this approach is easier? I.e. shouldnt we say that given the entitiy span the disanbiguation can be done by simple string matching etc.]}
The entity span detector can be implemented
based on classifiers as well, either as a pair of classifiers (one for predicting the start position and one for predicting the end position of the span) or as a collection of independent classifiers\footnote{representing a sequence tagger} (one for every position in the input, akin to a token classification network), where each classifier predicts whether the token at the corresponding position belongs to the entity span or not. 

%\dltodo{remove/dissolve the following}
%To conclude, 
To give a concrete example of a full KGQA semantic parsing algorithm relying on classification models,
we will describe the %\af{full} semantic parsing algorithm \afrm{and training procedur}
approach
proposed by 
\citet{mohammed2018}, %in more detail, 
chosen due to its simplicity and competitive performance on \dataset{SimpleQuestions}. 
%It leads to state-of-the-art results on \dataset{SimpleQuestions} and consists of the following steps:
This QA system follows the %following
following
steps for predicting the entity-relation pair constituting the formal query:

\begin{enumerate}
    \item Entity span detection: a bidirectional Long-Short Term Memory network (BiLSTM)~\citep{DBLP:journals/neco/HochreiterS97} is used for sequence tagging, i.e.~to identify the span $a$ of the NLQ \Q
    that mentions the entity.
    %\af{[Check: Did we introduce LSTMs before? Otherwise we need to do so here. And do we need the aberivation biLSTM?]}
        In the example ``\textit{Where was Michael Crichton born?}'', the entity span could be ``\textit{Michael Crichton}''. %The tagger is trained using annotations automatically generated by aligning questions with entity labels.
    
    \item Entity candidate generation:
    Given the entity span, all KG entities are selected whose labels (almost) exactly match the predicted span.
    %The resulting set of entities will further be pruned in the final step.
    
    \item Relation classification:
    A bidirectional gated recurrent unit (BiGRU)~\citep{DBLP:conf/emnlp/ChoMGBBSB14} encoder is used to encode $q$.
    %, which is produced by replacing the span $a$ in the original question $q$ with an entity placeholder (e.g.~with entity placeholder ``E'', the question pattern for our example query becomes``\textit{Where was} E \textit{born?}'').
    The final state of the BiGRU is used to compute probabilities over all relations using a softmax output layer, as described in Equations~\eqref{eq:af_trans} and \eqref{eq:softmax}.

    %(this is a basic classifier).
    %The most likely predicate that is also connected to any of the entities retrieved in Step 2 is selected as $r_{max}$ and is used in the predicted query.
    
    \item Logical form construction:
    The top-scoring entities and relations from previous steps are taken, and all their possible combinations are ranked based on their component scores and other heuristics. The top-scoring pair is then returned as the predicted logical form of the question.
    %given the relation $r_{max}$ predicted in Step 3, the set of entities from Step 2 is ranked according to the co-occurrence of the entities with relation $r_{max}$ in the KG. The top-ranked entity is used in the predicted query.
\end{enumerate}

%\afrm{This approach uses two models, a BiLSTM for entity span prediction and a simple BiGRU for relation classification.The training procedure consists of (1) generating data to train both models and (2) training the models.}
For training the full system, % the relation classifier, 
the relations given in the formal queries of the dataset are used as correct output classes. %in normal classifier training
For the span detector, we first need to extract ``pseudo-gold'' entity spans since they are not given as part of the dataset.
This is done by automatically aligning NLQs with the entity label(s) of the correct entity provided in the training data,
i.e.~the part of the NLQ that best matches the entity label is taken as the correct span to train the sequence tagging BiLSTM.

%\af{[TODO: Now related work could be discussed: paste simplequestions intro from ranking in here somewhere somehow and integrate what we need from the following seqgemnts:]]

\cite{mohammed2018} investigate different RNN and convolutional neural network (CNN) based relation detectors as well as bidirectional long-short-time-memory (LSTM)~\citep{DBLP:journals/neco/HochreiterS97} and CRF-based entity mention detectors. They show that a simple BiLSTM-based sequence tagger can be used for entity span prediction leading to results competitive to the state-of-the-art.

Both~\cite{petrochuk2018simplequestions} and \cite{mohammed2018} first identify the entity span, similarly to previous works, but disambiguate the entity without using neural networks.
% \citet{mohammed2018} show that a simple BiLSTM-based sequence tagger can be used for entity span prediction \af{leading to} results competitive \af{to} the state-of-the-art.
\citet{petrochuk2018simplequestions} employed a BiLSTM-CRF model for span prediction, which can no longer be considered a simple collection of independent classifiers and instead forms a structured prediction model that captures dependencies between different parts of the output formal query (in this case the binary labels for every position).

\subsection{Ranking based KGQA}
\label{sec:ranking}

If we cannot assume that all formal queries follow a fixed structure, as we could in the previous section, the task of mapping an NLQ to its logical form becomes more challenging. Since the set $\FQset$ of all possible formal queries grows exponentially with the length of the formal query,
%\ptrm{\nc{is too large because it grows exponentially with the length of the formal query,}} \ncrm{its too huge to be fully considered,} 
KGQA systems need to reduce the set of possible outputs. Ranking based semantic parsers therefore employ some kind of search procedure to find a suitable set of candidate formal queries $\mathcal{C}(\Q) =\{\FQ_1,...,\FQ_N\}$ for a given NLQ $\Q$ and the underlying KG, and then use a neural network based ranking model to find the best matching candidate.

\subsubsection{Ranking Models}
Formally, given an NLQ $\Q$ and a candidate set of formal queries $\mathcal{C}(\Q)$, the task of a ranking model is to output a score for each of the candidate formal queries in $\mathcal{C}(\Q)$ that allows to rank the set of candidate formal queries, where a higher score indicates a better fit for the given NLQ $\Q$.
%consist of three parts: an NLQ encoder, a formal query encoder,  and a scoring function
In a neural ranking setting, this is accomplished  via a two step process.
% \ptrm{In neural network based ranking models\nc{, as} a} 
In the first step, a neural % network based 
encoder model is %\ptrp{is}{s are} 
employed to map $\Q$ as well as every candidate $\FQ \in \mathcal{C}(\Q)$ is mapped to a %fixed-length 
latent representation space, resulting in the vector $\vec{\Q}$ for the NLQ and vectors  $\vec{\FQ}$ for every candidate.
%$\vec{\FQ}^{(1)}, \ldots, \vec{\FQ}^{(M)}$, respectively.
In the second step, each formal query encoding vector $\vec{\FQ}$, %$\vec{\FQ}^{(i)}$ $(i=1,\dots,M$), 
paired with the encoded question $\vec{\Q}$ is fed 
%\ptrm{\gm{In the second step, }the \pt{encoded question} %{NLQ embedding}
%$\vec{q}$ \ncrm{and} \nc{is paired with} each formal query \ptrp{encoding vector}{embedding} $\vec{\FQ_i}$ $(i=1,\dots,n$) \pt{which are together} \ptrm{\nc{and} }\ncrm{are then} fed} 
into a differentiable  scoring function, that returns a matching score $s\left(\Q, \FQ\right)$ indicating how well the formal query $\FQ$ matches the question \Q. The scoring function $s(\cdot,\cdot)$ can,
for example, be a parameterless function such as the dot product between the embedding vectors, or by a parameterized function, such as another neural network receiving the embeddings as input. The highest scoring formal query candidate is then returned as the prediction of the model:
%\nc{The inference task is thus a structured prediction task where the most plausible candidate formal query $\FQ^*$ is selected as follows:}
\begin{equation}
    \label{eqn:rankinference}
    \FQ^* = \argmax_{\FQ \in \mathcal{C}(\Q)} s_\theta\left(\Q, \FQ\right) \enspace.
\end{equation}
%where $s_\theta\left(\Q, \FQ\right)$
%is a differentiable function depending on the embeddings $\vec{q}$ and $\vec{f_i}$ that returns a matching score indicating how well the formal query $\FQ_i$ matches \Q; it can, for example, be computed by a parameterless function such as the dot product between the embedding vectors, or by a parameterized function, such as another neural network receiving the embeddings as input.
When multiple answers are possible for a given question, the following equation may be used to determine a set of outputs by also considering candidates whose scores are close to the best-scoring candidate~\citep{dong2015question}:
%\af{[Note: Are you sure there is a max in line (9)? If so its only the closest out of the close query. Dont we want all that are in gamma distance to the top ranked query.]}\ncnote{I have changed the $\FQ^*$ to \FQ to make it less ambiguous. The max() here finds the score of the highest scoring candidate (that we get in eq. 5): so we get all candidates that have a score within gamma distance of the best one (and also the best one}
%
\begin{align}
\label{eqn:rankinference_multi}
    \FQset_a^* = \lbrace \FQ \,|\, {}& \FQ \in \mathcal{C}(\Q) \;\; and \;\;
                        s_\theta(\Q, \FQ^{*}) - s_\theta(\Q, \FQ) < \gamma \rbrace \enspace.
\end{align}
% \begin{align}
% \label{eqn:rankinference_multi}
%     \FQset_a^* = \lbrace \FQ \,|\, {}& \FQ \in \mathcal{C}(\Q) \;\; and 
%                         \max_{\FQ^{'} \in  \mathcal{C}(\Q)} \lbrace s_\theta(\Q, \FQ^{'}) \rbrace - s_\theta(\Q, \FQ) < \gamma \rbrace
% \end{align}

% \af{[Note: We had some $f$ here before. Commented it out. Let me know if you believe we need it!]}
% \dltodo{changed eq labels because latex was breaking}
% \nc{[Note: Is this abusing argmax notation? Instead, should I do $k = \argmax_i blah$ say $L^k$ is the best candidate?]}

We can train ranking models using (i) \textit{full supervision}, where the training data consists of questions and corresponding logical forms, and (ii) \textit{weak supervision}, where gold logical forms are absent, and only pairs of questions and corresponding answers (i.e. execution results from the KG) are available for training. We discuss these two cases below.

% In order to train ranking models we need to distinguish two cases, as mentioned in Sec.~\ref{sec:nnkgqa}, namely - (i) \textit{Full Supervision} where the training data consists of questions, and corresponding logical forms, and (ii) \textit{Weak Supervision} where the training data does not contain logical forms corresponding to the question, but only the final answer. We discuss these two cases below.

\paragraph{Full Supervision} 
Given a dataset $\mathcal{D}^+ = \lbrace \left( \Q^{(i)}, \FQ^{(i)} \right) \rbrace_{i=1}^N$ of pairs of NLQs and their corresponding \emph{gold} formal queries,
% \af{(often referred to as \emph{golden}[Note: What is the standard name here?])},
a set of \emph{negative examples} $\mathcal{D}^- = \lbrace ( \Q^{(i)}, \hat{\FQ}^{(i)} ) \rbrace_{i=1}^N$ is generated. That is, for each NLQ $\Q^{(i)}$ in the training set, a false formal query $\hat{\FQ}^{(i)}$ that does not execute to the correct answer for $\Q^{(i)}$ is created. Typically, this process, referred to as \emph{negative sampling}, relies on a random or heuristic-based search procedure.% is employed  

% \nc{[Note: citation for negative sampling needed?]}
% \af{[Note: Is there a paper where different negative sampling strategies are explained?]}.
%
Training the ranking model then corresponds to fitting the parameters of the score function and the neural encoders %for the ranking model is learned 
by %optimizing 
minimising either a pointwise or pairwise ranking objective function.
% A \af{popular} pointwise objective \af{is} the logistic loss computed over individual positive and negative examples
% \begin{equation}
%     \label{eqn:logisticloss}
%     %\min_\theta
%     \sum_{\left(\Q, \FQ\right) \in \mathcal{D}^+} \log\left(1 + \exp(s_\theta\left(\Q, \FQ\right)\right) +
%     \sum_{\left(\Q, \hat{\FQ}\right) \in \mathcal{D}^-} \log\left(1 + \exp(-s_\theta\left(\Q, \FQ\right)\right) \enspace.
% \end{equation}
A popular ranking objective is the \textit{pairwise hinge loss} that can be computed over pairs of positive and negative examples:
\begin{equation}
    \label{eqn:hingeloss}
   % \min_\theta
    \sum_{\left(\Q, \FQ\right) \in \mathcal{D}^+} \sum_{(\Q, \hat{\FQ}) \in \mathcal{D}^-}
    \max (0, s_\theta(\Q, \hat{\FQ}) - s_\theta(\Q, \FQ) + \gamma) \enspace,
\end{equation}
Optimizing the model parameters $\theta$ by minimizing this function incentives the model to score  positive examples higher than %that of 
negative ones, where $\gamma > 0$ specifies the width of the margin between the scores of positive and negative examples.
Alternatively, ranking models can be trained in a pointwise manner by computing the logistic loss over individual positive and negative examples.

% \af{[Note: I am still not sure now, if it wouldnt be the best in the end to keep a sepearte weak supervision. This somehow breaks the readning flow here!.]}

\paragraph{Weak Supervision}
In order to train the ranking objectives under weak supervision, given a dataset of pairs of questions and corresponding execution answers $\mathcal{D}^+ = \lbrace \left( \Q^{(i)}, \A^{(i)} \right) \rbrace_{i=1}^N$, a search procedure is used to find \textit{pseudo gold} logical forms $\FQset_p^{(i)} = \lbrace \FQ | \FQ\left(\KG\right) = \A^{(i)}\rbrace$ for each question $\Q^{(i)}$ that evaluate to the correct answer $\A^{(i)}$. Pairs of questions and their corresponding \textit{pseudo gold} logical forms are then used to train the ranking models as described above.

The maximum margin reward (MMR) method introduced by~\citep{peng2017maximum} proposed a modified formulation of the pairwise hinge-loss based ranking objective in Eq.~\ref{eqn:hingeloss} called the most-violation margin objective. For each training example $\left( \Q, \A \right)$, they find the highest scoring logical form $\FQ^*$ from $\FQset_p$, as the \emph{reference} logical form. They define a reward-based margin function $\delta: \FQset \times \FQset \times \Aset \rightarrow \mathbb{R}$ and find the logical form that violates the margin the most:
\begin{equation}
    \label{eqn:mmrmostviolate}
    \hat{\FQ} = \argmax_{\FQ \in \FQset} \left( s_\theta(\Q, \FQ) - s_\theta(\Q, \FQ^*) + \delta(\FQ^*, \FQ, \A) \right)
\end{equation}
where $\delta(\FQ^*, \FQ, \A) = R(\FQ^*, \A) - R(\FQ, \A)$ , and $R(\FQ, \A)$ is a scalar-valued reward that can be computed by comparing the labelled answers \A and the answers $a^\prime = \FQ(\KG)$ yielded by executing \FQ on the KG. The reward function chosen by \cite{peng2017maximum} is the $F_1$ score. The most-violation margin objective for the training example $\left( \Q, \A \right)$ is thus defined as:
\begin{equation}
    \label{eqn:mmrhingeloss}
    \max (0, s_\theta(\Q, \hat{\FQ}) - s_\theta(\Q, \FQ^*) + \delta(\FQ^*, \hat{\FQ}, \A)) \enspace,
\end{equation}
% \pt{[PT: @NC we need to define the $R(f,a)$ function.]}
%\pt{[PT: Eq. 7 has the $\sum_a^b$ things signifying precisely what is sampled and how. Can we do something like that here in Eqn. 9 as well? Its not as straightforward since the $\hat{f}$s are generated dynamically so notations would be difficult though.]}
where $\hat{\FQ}$ is computed using Eq.~\ref{eqn:mmrmostviolate}. Note that the MMR method only updates the score of the reference logical form and the most-violating logical form. MMR essentially generalizes the pairwise hinge loss for non-binary reward functions.

Below, we discuss how existing approaches design the scoring function $s_\theta(\Q, \FQ)$, followed by a discussion on the parsing algorithms used for finding candidate logical forms in ranking based methods.

\paragraph{Encoding Methods and Score Functions}%\hfill\\ 
% \ncnote{The idea is to have a better structure by dividing the discussion into encoding formal queries vs searching/ranking them, like in the translation section.}
Neural network based models used for encoding questions and logical forms vary in complexity ranging from simple embedding based models to recurrent or convolutional models.
    
\cite{bordes2014question} proposed one of the first neural network based KGQA methods for answering questions corresponding to formal queries involving multiple entities and relations, and evaluate their method on \dataset{WebQuestions}. They represent questions as bag-of-words vectors of tokens in the question, and logical forms as bag-of-words vectors indicating whether a certain relation or entity is present in the logical form. These sparse bag-of-words representations are then encoded using a neural embedding model that computes the sum of the embeddings for the words, entities and relations present in the bag-of-words vector, yielding fixed-length vector representations for the  question and logical form respectively. Their match is scored by computing the dot product between the resulting vectors.

\cite{dong2015question} improve upon the simple embedding-based approach by introducing multi-column CNNs that produce three different vector-based representations of the NLQ. For each NLQ representation, they create three different vector representations for the logical form by encoding (i) the path of relations between the entity mentioned in the question and the answer entity, (ii) the 1-hop subgraph of entities and relations connected to the path, and (iii) the answer type information respectively, using embedding-based models similar to the above method. The sum of dot products between the question representations and their corresponding logical form representations is computed to get the final score.
% The use of structural information, such as entity type vectors, and pretrained KG embeddings{ A good introduction to such latent feature or KG embedding methods is given by \cite{nickel2016review}.} has also been shown to be beneficial for the KGQA task in \cite{dai2016}  and \cite{lukovnikov2017}. Note that these approaches focus on a subset of the task, namely answering \textit{simple questions} (See Sec.~\ref{}), and correspondingly train their models on \dataset{SimpleQuestions}. Instead of complex logical forms, approaches for this dataset require ....
% These, and other works typically encode the subject entity, and the relations in the logical form separately and rank them against the given NLQ, since ... 

\cite{zhang2016question} propose an attention-based representation for NLQs and incorporate structural information from the KG into their ranking-based KGQA model by embedding entities and relations using a pretrained TransE model~\citep{bordes2013translating}. They adopt a multi-task training strategy to alternatingly optimize the TransE and KGQA objectives respectively.

The use of structural information, such as entity type vectors, and pretrained KG embeddings\footnote{ A good introduction to such latent feature or KG embedding methods is given by \cite{nickel2016review}.} has also been shown to be beneficial for the KGQA task in \cite{dai2016}  and \cite{lukovnikov2017}. Furthermore, \cite{lukovnikov2017} explore building question representations on both word and character level.
% \pt{Both of these approaches are made to perform on \dataset{SimpleQuestions} task, and focus on answering only simple questions.}
Note that these approaches focus on a subset of the task, namely answering simple questions with a single entity and relation, and correspondingly train their models on \dataset{SimpleQuestions}.
    % \pt{As discussed in Sec.~\ref{}, the information required to answer simple questions includes one entity, and one relation.% As discussed in Sec.~\ref{}, answering these questions require the prediction of one entity and one relation}
These, and other works on simple questions typically encode the subject entity, and the relations in the logical form separately and rank them against the given NLQ, instead of treating them together as a formal query language expression. An exception to this is \citet{bordes2015large} which ranks entire triples. 

% \pt{Coming back to complex questions ... }
Instead of incorporating structural information from the KG, \citet{luo2018knowledge} incorporate local semantic features into the NLQ representation. To do so, they extract the dependency path between the entity mentioned in the NLQ and annotated \textit{wh-token}\footnote{``what'', ``where'', ``when'', ``who'', ``which'' or ``how''}, and encode both the dependency path as well as the NLQ tokens. %, thus incorporating local semantic features into the NLQ representation.

\citet{yih2015semantic} use CNNs to encode graph-structured logical forms called \emph{query graphs} after flattening the graph into a sequence of tokens. Their method is extended by \cite{yu2017improved} who use bidirectional LSTMs for encoding the logical forms. They use two kinds of embeddings for encoding relations: relation-specific vector representations, as well as  word-level vector representations for the tokens in the relation.

The approaches discussed above are implemented over the \dataset{Freebase} knowledge graph. 
% \ncnote{Citing to-be-published ISWC paper - check if bibtex is ok}
\cite{maheshwari2018learning} propose an attention-based method to compute different representations of the NLQ for each relation in the logical form, and evaluate their approach on \dataset{LC-QuAD} and \dataset{QALD-7}.
% \af{[Todo: We should put a reference to the ICSW version of the paper :) ]}
They also show that transfer learning across KGQA datasets is an effective method of offsetting the general lack of training data, by pre-training their models on \dataset{LC-QuAD}, and fine-tuning on \dataset{QALD}.
Additionally, their work demonstrates that the use of pre-trained language models~\citep{DBLP:conf/naacl/DevlinCLT19, radford2019language, DBLP:journals/corr/abs-1906-08237} for KGQA is a potentially beneficial technique for further increasing the model's performance.
\citet{lukovnikov2019} also explores the use of pre-trained language models for the KGQA task, over \dataset{SimpleQuestions} (which uses \dataset{Freebase}).
\subsubsection{Ranking based Parsing Algorithms}
\label{sec:ranking:parsing}
As mentioned above, searching for candidate logical forms is a pivotal step in training and inference of ranking-based parsing algorithms.
During inference, a search procedure is used to create the formal query candidate set $\mathcal{C}(\Q)$, which is then ranked using the scoring function.
The size of the search space crucially depends on the complexity of the NLQs the systems aims to answer.
In the weakly supervised setting, in the absence of gold annotated formal queries, an additional search procedure is required during training to \textit{guess} the latent formal queries that resolve to the correct answer for a given NLQ.
A similar search process may be used for generating the negative examples needed for training ranking models.

Typically, the search procedure builds on multiple techniques to restrict the candidate space, such as:
 \begin{itemize}
     \item [(i)] using off-the-shelf entity-linking tools to generate entity candidates, and limiting the search space to formal queries containing only entities and relations found within a certain distance (usually measured in number of relation traversals, also referred to as \textit{hops}) of the candidate entities,
\item[(ii)] limiting the number of relations used in the formal query,
\item[(iii)] employing \textit{beam search},  a heuristic search algorithm that maintains a \textit{beam} of size $K$, i.e. a set of at most $K$ incomplete output sequences that have high probability under the current model; at every time step during prediction, every sequence in the beam is expanded with all possible next actions or tokens, and the probabilities of the expanded sequences are evaluated by the scoring function, following which only the $K$ most likely sequences are retained in the beam and considered for further exploration,
\item[(iv)] strict pruning of invalid or redundant logical forms that violate grammar constraints of the formal language, or those that do not adhere to the structure of the KG, along with additional rules and heuristics.
\end{itemize}

\cite{bordes2014question} and \cite{dong2015question} employ (i) and (iii), i.e. they find the entities mentioned in the question, and use a beam search mechanism to iteratively build logical forms one prediction at a time, using their ranking model. 
% {\cite{bordes2014question} and \cite{dong2015question} employ beam search over their ranking models to map a question to its correct answers, by generating logical forms in an end-to-end manner.}
In contrast, \cite{xu2016question} propose a multi-stage approach in order to better handle questions with multiple constraints. They use syntactical rules to break a complex question down into multiple simple questions, and use multi-channel CNNs to jointly predict entity and relation candidates in the simple fact-based questions. The predictions of the CNN are validated by textual evidence extracted from unstructured Wikipedia pages of the respective entities. They train an SVM rank classifier \citep{joachims2006training} to choose the best entity-relation pair for each simple question.

\citet{yih2015semantic} employ a heuristic reward function to generate logical forms using a multi-step procedure, starting by identifying the entities mentioned in the NLQ, then building a \textit{core chain} consisting only of relations that lead to the answer entity, and finally adding constraints like aggregation operators to their logical form.
Their logical forms, called \textit{query graphs}, can be viewed as the syntax tree of a formal query in $\lambda$-DCS\footnote{Concretely, the core chain can be described as a series of conjunctions in $\lambda$-DCS, where the peripheral paths of the tree are combined using intersection operators.}~\citep{liang2013lambda}.
For a given NLQ, partial query graphs generated at each stage of the search procedure are flattened into a sequence of tokens. The NLQ and the formal query tokens are respectively encoded by a CNN model and the dot product of the output representations is used as the final output score.

\label{sec:complexquestions}

\subsection{Translation based  KGQA}
\label{sec:translation}
%\af{[Note:Can we distinguish here in simple and complicated questions as well (To follow the same structure as above)?]}

%Question answering over knowledge graphs (or other structured data) is generally accomplished by first translating the question into a formal query (logical form) that can be executed over the knowledge source. In the case of knowledge graphs like DBpedia, Freebase or Wikidata, the formal query can be translated into SPARQL. For tabular data, the formal query can be written in SQL. 
%The task of mapping natural language expressions to an executable formal representation is known as \textit{semantic parsing}.
%The previously discussed approaches perform question answering by learning to choose the correct logical form among a (large) set of pre-generated candidates.
In this section, we will focus on methods that learn to \emph{generate} a sequence of tokens that forms the logical form as opposed to learning to choose the correct logical form among a set of pre-generated candidates.

Semantic parsing in this setup is modelled as a translation problem, where we need to translate a NLQ \Q into a formal query $\FQ$ that can be expected to return the intended answer when executed over the source KG \KG.
A popular approach for mapping sequences in one language to sequences in another language is to use neural sequence-to-sequence (seq2seq) models.
Such models were first introduced for machine translation, i.e., for mapping a sentence in one natural language to the corresponding sentences in another~\citep{bahdanau,luong}. With some changes, this neural architecture has been extensively adapted for semantic parsing~\citep{dong2016,jia2016,krishnamurthy2017,seq2sql,sqlnet,stamp} and more specifically for the KGQA task~\citep{golub2016,nsm,structvae,cheng2018Weakly,guo2018dialog}.

\subsubsection{Translation Models}

A typical neural sequence-to-sequence model consists of an encoder, a decoder and an attention mechanism. The encoder encodes the input sequence to create context-dependent representations for every token in the input sequence. The decoder generates the output sequence, one token at a time, conditioning the generation process on previously generated tokens as well as the input sequence. The attention mechanism~\citep{bahdanau,luong} models the alignment between input and output sequences which has proven to be a very useful inductive bias for translation models. %A more formal description of sequence-to-sequence models is provided in Section TODO.
In neural sequence-to-sequence models, generally RNNs are used to encode the input and to predict the output tokens. % is typically generated using a recurrent neural network (RNN) generator.
%Even though an RNN is the most common choice, o
Other encoder/decoder network choices are also possible, such as CNNs and transformers~\citep{transformer}.
Since logical forms are generally tree-structured, and a basic sequence decoder does not explicitly exploit the tree dependencies, several works have focused on developing more structured decoders (see Section~\ref{sec:treedecoder}).
% \dltodo{add picture}

Formally, given an input sequence  $\Q_{0\dots T}$\footnote{We use the notation $\Q_{0\dots T}$ to denote $(\Q_0, \dots ,\Q_T)$, the sequence of symbols $\Q_i$, for $i \in [0, T]$} of $T$ tokens from the input vocabulary $\mathcal{V}^I$ and an output sequence $\FQ_{0\dots T^*}$ of $T^*$ tokens from the output vocabulary $\mathcal{V}^O$, the translation model with parameters $\theta$, models the probability $p_{\theta}(\FQ_{0 \dots T^*} | \Q_{0\dots T})$ of generating the whole output sequence given the input sequence.
% \af{[Note: wouldnt it be better to denote the output sequence by $f$ instead of $y$ since we are talking about the formal query and we used $f$ to descibe it before.]}
The probability of the whole output sequence % $y_{0 \dots N}$
can be decomposed into the product of probabilities over the generated tokens
\begin{equation}
\label{eq:seqprob}
    p_{\theta}(\FQ_{0\dots T^*} | \Q_{0\dots T}) = \prod_{j=0}^{T^*} p_{\theta}(\FQ_j | \FQ_{<j}, \Q_{0 \dots T}) \enspace ,   
\end{equation}
where %$x_{0..T}$ is the input sequence of $T$ tokens from the input vocabulary $\mathcal{V}^I$, $y_{0..N}$ is the output sequence of $N$ tokens from the output vocabulary $\mathcal{V}^O$ and
$\FQ_{<j}$ is the sequence of tokens generated so far, \af{i.e.~}$\FQ_{<j}=(\FQ_0, \dots ,\FQ_{j-1})$. 
%\af{[Note: We should check if we want to keep this sequence notation. It might be a bit unsual and therefore maybe better be explained]}

During prediction, the output sequences are usually generated by taking the most likely token at every time step, which is also referred to as \emph{greedy search}, that is
\begin{equation}
    \FQ_j = \argmax_{\FQ \in \mathcal{V}^O} p_{\theta}(\FQ | \FQ_{<j}, \Q_{0..T})
    \enspace,
\end{equation}
or by using \emph{beam search} (see Section.~\ref{sec:ranking:parsing}). 
%We discuss various methods to constrain the space of tokens at every time step in Section~\ref{sec:translation:parsing}.

\begin{comment}
%Rather than taking the locally best token at every time step,
\pt{[PT:should remove the rest of this para.]}
Beam search \af{is a heuristic search algorithm that}  maintains a set of \af{the highest probably} incomplete sequences \af{at each time step} %with the highest probability %(as in Eq.~\ref{eq:seqprob}) 
 \af{performing a} search with multiple candidates which \af{increases} the chance of finding the most probable sequence. At every time step, every sequence in the \af{set} is expanded with all possible next tokens, the probabilities of the expanded sequences are evaluated, and only the top-$K$ sequences (\af{where} $K$ is \af{referred to as} beam size) are retained in the \af{set of sequences considered in} the next time step.
%\af{[Note: Can we add a short explaination here?]}
\end{comment}

In the following, we discuss common ways to train translation based KGQA models in the fully and weakly supervised settings.

\paragraph{Full Supervision}
In the \textit{fully supervised} setting, the parameters $\theta$ of the model $p_{\theta}(\FQ_j | \FQ_{<j}, \Q_{0..T})$ are typically trained by maximizing  their likelihood given the %observed 
sequence pairs in the training set $\mathcal{D} =
\left\{ \left( \Q^{(i)}, \FQ^{(i)} \right) \right\}_{i=1}^N$, that is, by maximizing the following objective:
\begin{equation}
    %\theta^* = \argmax_{\theta} 
    %\obj_{ML} = 
    \sum_{(\Q, \FQ) \in \mathcal{D}}p_{\theta}(\FQ_{0\dots T^*} | \Q_{0\dots T}) \enspace .
\end{equation}

%\ptrm{\af{where the expectation is taken over the empirical distribution.
%[Note: We dont have an expectation explicitly in the qeuation anymore!]}}
%\afrm{The maximum likelihood  objective \af{is used when} full supervision (questions and their queries) are provided.}

\paragraph{Weak Supervision}
%\paragraph{Weak supervision:}
% \af{[Note: Integrate the following!]}
%\nc{[Note: This section will be removed. The algorithms stuff is being integrated above into Learning and Inference, details on weakly supervised works will be integrated into the approaches sections wherever they fit.]}
\label{sec:weaksup}

In the \textit{weakly supervised} setting, the training data $\mathcal{D} = \lbrace \left( \Q^{(i)}, \A^{(i)} \right) \rbrace_{i=1}^N$ consists only of pairs of NLQs and corresponding execution results.
Different methods have been proposed for training translation models for semantic parsers using only weak supervision.
Even though the proposed methods vary in the objective function and training procedure, they all operate on the general principle of maximizing the probability of producing the correct execution results.
%Weakly supervised question answering approaches vary widely \af{in the learning principle they apply and in} the kind of objective function being \af{used} (e.g. reinforcement learning, iterative maximum likelihood, marginal maximum likelihood) and the way the logical forms and execution results are generated (explicit semantic parsing with discrete, non-differentiable execution vs. smooth, differentiable execution).

%\af{[Note/Todo: We dont use names for losses/training objectives so far, so we dont need to do it now!]}\pt{[agreed!]}
%\pt{Some of the common ways to accomplish this}
%\af{In} the following  common ways to train a \afrm{semantic parsing model for question answering} \af{translation model} under weak supervision \af{are listed}:
The following training methods are commonly used for weakly supervising translation-based semantic parsing models: %{have been proposed in literature}:
\begin{enumerate}
    \item \textbf{Maximum Marginal Likelihood (MML):} 
    %\afrm{With access to the correct logical forms (fully supervised setting), we can use the maximum likelihood (ML) objective to try to maximize the probability of producing the correct logical form.}
    The maximum marginal likelihood method follows a similar %formulation 
    idea as the maximum likelihood (ML) method used in the fully supervised setting, but instead of directly maximizing the probability of producing the correct logical form, it maximizes the probability of producing the correct execution results.
    In order to do so, MML marginalizes over all possible logical forms, maximizing the following objective function:
    \begin{equation}
        \label{eq:mml:1}
       % \mathcal{O}_{MML} = 
       \sum_{(\Q, \A) \in \mathcal{D}} \log p(\A | \Q) = \sum_{(\Q, \A) \in \mathcal{D}} \log \sum_{\FQ \in \FQset} p(\A | \FQ) p_{\theta}(\FQ | \Q) \enspace ,
    \end{equation}
    where $\A$ is the correct answer to the NLQ $\Q$ and the  sum is computed over all \FQ's in the space of all possible logical forms \FQset.
    Since semantic parsing environments are usually deterministic (i.e.~the same query always executes to the same answer when the KG is kept fixed), the $p(\A | \FQ)$ term is either $1$ (if \FQ's execution produced the correct results $\A$, i.e. $\FQ(\KG) = \A$) or $0$, which leads to the following simplified notation
    \begin{equation}
        \label{eq:mml:2}
        %\mathcal{O}_{MML} = 
        \sum_{(\Q, \A) \in \mathcal{D}} \log \sum_{\FQ \in \FQset^*} p_{\theta}(\FQ | \Q) \enspace ,
    \end{equation}
    where $\FQset^*$ is the set of consistent logical forms that execute to the correct answer $a^{(i)}$.
    %
    %Formally assumes that $y$ is generated by a partially-observed random process: conditioned on the question $x$, a latent program $\mathbf{z}$ is generated \af{[Note: what is latent program? A latent variable? And is it really sampled or deterministically generated?]}, and conditioned on $\mathbf{z}$, the observed answer $y$ is generated~\citep{guu2017bridging}. This implies the marginal likelihood:
    %\begin{equation*}
    %    p_{\theta}(y | x) = \sum_{\mathbf{z} \in \mathcal{C}(x, y)} p(y | \mathbf{z}) p_{\theta} (\mathbf{z} | x)
    %\end{equation*}
    %\af{[Note: Shouldnt we have a conditional instead of the joint of y and z?]}
    %\dl{[Yes, fixed]}
    %and the objective function is therefore, $\mathcal{O}_{IML}(\theta) = \sum\limits_{(x,y)} \log p_{\theta}(y | x)$.
    %\dl{Note that $p(y | \mathbf{z})$ is 1 if the logical form $z$ leads to the correct answer $y$ and 0 otherwise and thus can be thought of as reward. [TODO: someone check this]}
    %\af{[Note: if we give the sum over the training set here wouldnt we need it as well in tge IML objective?]}\dl{[yes, added sum above]}
    %
    The set $\FQset^*$ is usually approximated using online beam search.
    %\afrm{in $\FQset^*$ and used according to Eq.~\ref{eq:mml:1}}. 
    However, an approximation of $\FQset^*$ can also be computed beforehand~\citep{dpd} and kept fixed throughout training.
    %\af{[Todo: fix citation!]}
    %MML with beam search is traditionally used to learn semantic parsers from weak supervision - for each question, a set of \textit{pseudogold} logical forms are constructed and their marginal likelihood is optimized. Since there is no supervision on logical forms, those candidates that evaluate to the ground truth answer are considered \textit{pseudogold}.
    
    \item \textbf{Reinforcement Learning using Expected Rewards (ER):} 
    The problem of weakly supervised semantic parsing can also be approached as a reinforcement learning (RL) problem.
    In fact, we can view the translation model as a parameterized policy that given a state, which in our case consists of the KG \KG, input \Q and decoding history $\FQ_{<t}$,
    % \af{[Note: It is unclear what $\FQ_{<t}$ is referring to! An partially decoded formal query? Do we need to introduce the notation, i.e. is it used somewhere else?]}
    must decide what action to take in order to maximize a reward. In our case, the reward function $R(\FQ, \A)$ can be a binary reward at the end of an episode (end of decoding phase), that is $1$ if the produced trajectory (logical form) \FQ executes to the correct answer $\A$ and $0$ otherwise.
    The RL setup for semantic parsing is characterized by (1) a deterministic environment (the next state produced by executing the action on the current state is always the same) and (2) sparse rewards (the reward is given only at the end of the episode and, given the huge number of possible logical forms, is likely to be 0 most of the time).
    % \dltodo{Someone check this}.
    In addition, weakly supervised training of semantic parsing is characterized by \textit{underspecified} rewards~\citep{agarwal2019LearningTG} which could lead to learning \textit{spurious} logical forms. However, the reward function based on execution results only cannot take this into account.
    
    \textit{Policy gradient} methods are trained by optimizing the \emph{expected reward}:
     \begin{equation}
        \label{eq:er}
         %\mathcal{O}_{ER} = 
         \sum_{(\Q, \A) \in \mathcal{D}} \sum\limits_{\FQ \in \FQset} R(\FQ, \A) p_{\theta}(\FQ | \Q) \enspace ,
     \end{equation}
    %From the perspective of reinforcement learning (RL), $p_{\theta}(\FQ | \Q)$ is a parameterized policy function that generates a sequence of decisions $\FQ$ (a logical form), and then receive a reward at the end of the episode: $R(\FQ) = 1$ if $\FQ(\mathcal{K}) = a_i$ and $0$ otherwise. \textit{Policy gradient} methods solve the problem by optimizing the expected reward \af{for} a given example $(x,y)$:
    %\begin{equation*}
    %    G(x,y) = \sum_{\mathbf{z}} R(\mathbf{z}) p_{\theta}(\mathbf{z} | x)
    %\end{equation*}
    where the sum is over all possible logical forms, which, in practice, is estimated using an approximation strategy such as \textit{Monte Carlo integration}, i.e.~based on trajectories sampled from the policy.

    \item \textbf{Iterative Maximum Likelihood (IML):} The Iterative Maximum Likelihood (IML) objective \citep{nsm} uniformly maximizes the probability of decoding \textit{all} consistent logical forms across all examples: 
    \begin{equation}
    \label{eq:iml}
   % \mathcal{O}_{IML}(\theta) = 
   \sum_{(\Q, \A) \in \mathcal{D}} \sum\limits_{\FQ \in \FQset^*} \log \ p_{\theta}(\FQ | \Q) \enspace ,
    \end{equation}
    %\af{[Note: Shouldnt C depent on $y$ rather than on $x$? And do we actually need $\mathcal{O}_{IML}$? otherwise we might not need the notation.]} \dl{[I think both]}
    %is a \dl{``buffer''} set of candidate logical forms that have been found so far for $x$ that \dl{execute to the correct answer ($\mathbf{z}(\mathcal{K}) = y$)}.
    %\dl{[TODO: how is buffer populated]}
% \af{[Note: What does the followong sentecne is trying top say?]}\pt{{PT@AF: see my suggested change.}}
    When training policy gradient methods using ER, ~\citep{nsm} demonstrate that better exploration can be achieved by employing IML to populate an initial buffer of diverse trajectories (i.e. logical forms) which are then used to pretrain the RL model.

    \item \textbf{MAPO:} In order to reduce the variance of the ER estimator, \citet{liang2018memory} proposed a novel method that integrates a memory buffer.
    They reformulated the ER objective as a sum of action sequences in a memory buffer $\mathcal{C}(\Q)$ and outside the buffer:
    \begin{align}
        % \obj_{MAPO}() &= 
        \sum_{(\Q, \A) \in \mathcal{D}} \sum_{\FQ \in \mathcal{C}(\Q)} R(\FQ, \A) p_{\theta} (\FQ | \Q) + \sum_{\FQ \notin \mathcal{C}(\Q)} R(\FQ, \A) p_{\theta} (\FQ | \Q)     \enspace .
    \end{align}
    The memory buffers $\mathcal{C}(\Q)$ for each example $i$ are populated using \textit{systematic exploration}, which prevents revisiting sequences that have already been explored.
    %MAPO is derived from the ER objective and in the presented form leads to an unbiased low-variance estimator.
    When the two terms are explored further (see~\cite{liang2018memory}), we can see that the weight assigned to trajectories from the memory buffer is low in the beginning of training, when the policy still assigns low probabilities to the buffered trajectories.
    In order to speed up training, \citet{liang2018memory} propose \textit{memory weight clipping}, which amounts to assigning the buffered trajectories an importance weight of at least a certain value.
    Experimental results of \citet{liang2018memory} show significant improvement of the proposed MAPO procedure against common baselines %, including MML, ER and IML 
    and show that both systematic exploration and memory weight clipping are essential to achieve high performance.
    %MAPO fails when either systematic exploration or memory weight clipping are not used.
    
    %Training with this objective ensures that high-reward action sequences are included
    %\dltodo{explain what it does and how it helps}
    
    \citet{agarwal2019LearningTG} argue that apart from an optional entropy term, MAPO does not encourage exploration, which can be problematic. They propose a MAPO variant called MAPOX where the initial buffer is additionally populated by sequences found using IML. 
    %\dl{They also explore meta-learning of an auxiliary reward function that is used to decrease the effect of spurious logical forms during training.}
\end{enumerate}

%\dltodo{spurious programs: generative ranker and MeRL~\cite{agarwal2019LearningTG}}

%\dltodo{Move to ranking weak sup?}
%Methods like \citep{bordes2014question, yih2015semantic} discussed in previous sections rely on manual feature engineering to generate candidate logical forms which are ranked by embedding based models. STAGG~\citep{yih2015semantic} is a semantic parsing based method that \textit{grows} candidate logical forms in manually defined stages of increqasing complexity, and finally ranks them. 

%\afrm{\paragraph{Tackling spurious logical forms:}}
%\af{[Note: We talk also about the different objectives not only spurous forms so the prgraph name seems not to fit!]}
A disadvantage of the RL and MML approaches,
%\af{[Note: Are all except from the MLL objective regarded as RL? IML seems not to be!]} -- yes, this sentence only applies to RL and MML
noted by \cite{guu2017bridging} and \cite{agarwal2019LearningTG}, is that the exploration of trajectories is guided by the current model policy, which means that logical forms that have high probability under the current model are more likely to be explored. Consequently, logical forms with low probability may be overlooked by exploration.
%\ncrm{, once a high-reward trajectory is discovered, the policy will settle on this trajectory, thus possibly ignoring other high-reward trajectories.}
    In the presence of many spurious logical forms and only a few correct logical forms, it becomes more likely that spurious ones are explored first. Once the policy settles on these high-reward (but spurious) logical forms, the exploration gets increasingly biased towards them during training, leading to poor generalization.
   % \ncrm{, this may mean that the policy is more likely to settle on spurious logical forms during training and generalize poorly.}
    One common solution to this
    %\dlrm{\af{problem} %\afrm{used in RL}} 
    is to perform $\epsilon$-greedy exploration, which \citet{guu2017bridging} extend from RL to MML.
    %\af{[Note: Where is the following half sentence belonging to:]}
    \citet{guu2017bridging} also propose a
    \textit{meritocratic update rule} that updates parameters such that probability is spread more evenly across consistent logical forms. %\dl{In the limit of the smoothing parameter, the adapted MML objective becomes equivalent to the IML objective.}
    %The overall expected reward is $\mathcal{O}_{RL}(\theta) = \sum\limits_{(x,y)} G(x,y)$.
    %\dl{\citet{guu2017bridging} further discusses the relationship between RL and MML for weakly supervised semantic parsing.}    
    %While MML maximizes the product of marginal probabilities, expected reward maximizes the sum.\af{[Note: We have a sum in the formula above as well! Is that a typo?]} %More discussion of these two objectives can be found in 
    For a more detailed discussion of the MML and ER objectives, we refer the reader to the work of
    \citet{guu2017bridging}, \citet{norouzi2016reward}, \citet{roux2016tighter} and \citet{misra2018policy}.
    
\citet{agarwal2019LearningTG} proposes a meta-learning approach for learning an auxiliary reward function that is used to decrease the effect of spurious logical forms in the MAPOX objective.
The auxiliary reward function is trained such that the update to the policy under the reward augmented by the auxiliary reward function results in better generalization over a held-out batch of data.

%\af{[Note: In the followng it is hard to understand why we suddenly speeak about ranking models again!]} -- the rankers used here are proposed as a way of filtering out spurious logical forms for weakly supervised translation-based semantic parsers
\citet{cheng2018Weakly} propose a neural parser-ranker based approach for weakly supervised semantic parsing, where a sequence-to-sequence \textit{parsing} model (trained with beam search and IML) is coupled with a basic ranking model (trained using MML) as well as an \textit{inverse parsing} model, which is a generative model that reconstructs the original question from the logical form generated by the \textit{parsing model}. The reconstruction loss is used to further refine the parsing model in order to tackle the problem of spurious logical forms. Note that here the logical form is treated as a latent variable.
%\af{[Note: Here the distingtion between generative and discrimintative approaches seems to be less claer!]}

% \pt{[PT: I think we should remove this small paragraph for now. We can continue with the next para without this.]}Below follows a review of several works proposing interesting extensions or \textit{alternative training methods} to the common ones discussed above.
% }{ just covered}.
% \dltodo{Might be problematic to have the following as-is in latest structure.}
% \af{[Note: It might better fit in the tranlation based parsing section!]}

Another class of neural approaches for KGQA with weak supervision perform multi-step knowledge base reasoning in a fully differentiable end-to-end manner. TensorLog is a recently-developed differentiable logic that performs approximate first order logic inference through a sequence of differentiable numerical operations on matrices. NeuralLP~\citep{yang2017differentiable}, inspired by TensorLog, learns to map an input question to its answer by performing multi-step knowledge base reasoning by means of differentiable graph traversal operations.

The neural programmer~\citep{neelakantan2015neural} is a fully differentiable encoder-decoder based architecture augmented with a set of manually defined discrete operators (e.g. argmax, count), which is applied to the task of table-based complex question answering by \citet{neelakantan2016learning}. The discrete operators allow the model to induce \textit{latent logical forms} by composing arithmetic and logical operations in a differentiable manner. The model is trained using a weak supervision signal which is the result of the execution of the correct program.

In contrast, the Neural Symbolic Machines framework~\citep{nsm} combines (i) a differentiable neural \textit{programmer}, which is a seq2seq model augmented by a key-variable memory that can translate a natural language utterance to a program, and (ii) a symbolic \textit{computer} (specifically, a Lisp interpreter) that implements a domain-specific language with built-in functions and code-assistance that is used to prune syntactically or semantically invalid canddiate logical forms and execute them to retrieve a result that is used to compute the supervision signal. REINFORCE, augmented with \textit{iterative maximum likelihood training}, is used to optimize for rewarding programs. \citet{liang2018memory} demonstrate the effectiveness of a memory buffer of promising programs, coupled with techniques to stabilize and scale up training as well as reduce variance and bias.

%\citeauthor{guu2017bridging} \citeyear{guu2017bridging} argue that the deterministic beam search generally used for MML suffers from biased exploration in the presence of spurious programs. Combining the strengths of two common learning paradigms, RL and maximum marginal likelihood, the authors propose an $\epsilon$-\textit{greedy randomized beam search} to tackle this problem, as well as a \textit{meritocratic update rule} that updates parameters such that probability is spread more evenly across consistent logical forms.

\paragraph{Model Variations}
Several extensions have been proposed to the standard sequence-to-sequence model that try to improve the translation model for semantic parsing in general, as well as some more task- or dataset-specific extensions.

\label{sec:treedecoder}
% PT -> DL: Dont need intro here, last para above sounds like a good intro (see above yellow text)
%Several works on semantic parsing explored tree-structured decoders, which allow to model the hierarchical tree structure of queries more naturally than sequence decoders.
Semantic parsers with \textbf{structured decoders} use the same sequence encoders to encode the NLQ but induce additional structure on top of the normal attention-based sequence decoder that exploits the hierarchical tree structure of the query.
%\dltodo{something more?}

%The logical forms that need to be generated typically follow a tree structure that reflects the predicate argument structure or the parse tree of a programming language. However, since a sequence-to-sequence model (see Section~\ref{sec:seq2seq:intro}) generates sequences, and requires remembering all useful history in its internal RNN state, it may be less suited for decoding tree structures.
%A normal sequence decoder places the burden of generating correctly structured trees (e.g. matching parentheses) onto the memory of the decoder. In addition, a sequence decoder's states do not follow the structure of the tree, where it might be beneficial for the children states to have direct access to the parent state.
%For these reasons, several works explored tree-structured decoders, decoders that model dependencies according to the tree structure of the queries.

\citet{dong2016} %\citeyear{dong2016}
propose a tree decoding model that decodes a query tree in a top-down breadth-first order. 
For example, instead of decoding the lambda-calculus logical form \texttt{(argmin \$0 (state:t \$0) (size:i \$0))} (corresponding to the question ``\textit{Which state is the smallest?}'') as a sequence, the decoder first decodes the topmost level of the query tree \texttt{(argmin \$0 \textless n\textgreater \textless n\textgreater \textless /s\textgreater)}. Here, \texttt{\textless n\textgreater} and \texttt{\textless /s \textgreater} are artificial topological tokens introduced to indicate the tree structure: \texttt{\textless n\textgreater} is a non-terminal token indicating that a subtree is expected at its position and \texttt{\textless /s\textgreater} is the end-of-sequence token indicating the end of a sequence of siblings. After decoding this top-level sequence, the two children subtrees, \texttt{(state:t \$0)} and \texttt{(size:i \$0)}, are decoded by conditioning them on the the two non-terminal states. 
%The decoding of the child subtree is directly conditioned on the parent state (at the corresponding non-terminal) by passing the parent state as an input at every decoding step.
%
%Compared to a normal depth-first sequence decoder, in this breadth-first tree decoder the sibling subtrees can not be conditioned on each other since the trees are decoded in a breadth-first order and the state of the decoder after it's done decoding a child sequence is simply discarded.
%\af{[Note: the last sentence seems to be in some sense recursive and is not very clear.]}
From experimental results on \dataset{Geo880}, \dataset{ATIS}, \dataset{JobQueries} and \dataset{IFTTT}, it appears that the inductive bias introduced by the tree decoder improves generalization. 
%The improvement w.r.t. a sequence-to-sequence model is more noticable in smaller datasets (\dataset{Geo880} and \dataset{JobQueries}), than in the bigger ones (\dataset{ATIS} and \dataset{IFTTT}).
%, the improvement could be argued to be negligible. A possible explanation is that the previously mentioned breaking of dependencies between sibling subtrees reduces overfitting, which is easier to elicit  with smaller datasets.

\citet{alvarez2017}
propose an improved tree decoder, where the parent-to-child and sibling-to-sibling information flows are modeled with two separate RNNs. With this model, each node has a parent state and a previous-sibling state, both of which are used to predict the node symbol and topological decisions.
Instead of modeling topological decisions (i.e.~whether a node has children or further siblings) through artificial topological tokens like \citet{dong2016}, %\citep{alvarez2017} instead
they use  auxiliary classifiers at every time step that predict whether a node has children and whether the node is the last of its siblings.
%employed at every time step. 
The elimination of artificial topological tokens reduces the length of the generated sequences, which should lead to fewer errors. 
The decoding proceeds in a top-down breadth-first fashion, similarly to \citet{dong2016}.
%, which also prevents siblings subtrees from conditioning on each other. 
Experimental results on the \dataset{IFTTT} semantic parsing dataset show improvement obtained by the introduced changes.

%\dltodo{https://arxiv.org/pdf/1511.00060.pdf}

\citeauthor{cheng2017} (\citeyear{cheng2017,cheng2018Weakly,cheng2019learning})
%, \citet{cheng2018Weakly} and \citet{cheng2019learning}
develop a transition-based neural semantic parser that adapts
the Stack-LSTM proposed by \citet{dyer2015}.
%\afrm{(using \dataset{WebQuestions} and \dataset{GraphQuestions})}. 
The Stack-LSTM decodes the logical forms in a depth-first order, decoding a subtree completely before moving on to its siblings. 
The Stack-LSTM of \citet{cheng2018Weakly} uses an adapted LSTM update when a subtree has been completed: %performs normal LSTM updates when producing siblings or parent nodes. However, when it is indicated that a subtree has been finalized, 
it backtracks to the parent state of the completed subtree, 
%using this state for the next LSTM update, 
and computes a summary encoding of the completed subtree and uses it as input in the next LSTM update.
\citet{cheng2019learning} also shows how to perform bottom-up transition-based semantic parsing using the Stack-LSTM.
%, instead of feeding the embedding of the last generated token.
%This allows the Stack-LSTM to model dependencies between two sibling subtrees, unlike the two tree decoders discussed above. 
\citet{cheng2017} obtains state-of-the-art results on \dataset{GraphQuestions} and \dataset{Spades} and results competitive with previous works on \dataset{WebQuestions} and \dataset{GeoQueries}. 
\citet{cheng2018Weakly} further improves performance of their Stack-LSTM model on the weakly supervised \dataset{GraphQuestions}, \dataset{Spades} and \dataset{WebQuestions} datasets by using a generative ranker.

\textbf{Copying Mechanisms}~\citep{pointernet,seecopy,gucopy,jia2016} are an  augmentation of the seq2seq neural architecture which enables a direct copy of tokens or sub-sequences from the input sequences to the output.
Though it~\citep{pointernet,seecopy,gucopy,jia2016} is not generally required for semantic parsing, it can be useful for certain tasks or datasets.
In \dataset{WikiSQL}, for example, a copying mechanism is required to copy SQL condition values into the query~\citep{incsql,sqlnet,seq2sql}.
It has also been used for semantic parsing in general~\citep{jia2016,damonte2019practical}. % \pt{specifically in dealing with rare words such as uncommon entity mentions in the question}.

Another important aspect is \textbf{symbol representation}. When answering questions over large scale knowledge graphs, we are confronted with a large number of entities and relations not present in the training data. An important challenge, thus, is to find a way to learn representations for entities and relations that generalize well to unseen ones.
Representing both entities and predicates as a sequence of words, characters or sub-word units (BPE/WordPiece) in their \textit{surface forms}, as opposed to arbitrary symbols unique to each entity or predicate, offers an extent of generalizability to these unseen or rare symbols. %~\citep{yu2017improved,lukovnikov2017,maheshwari2018learning,golub2016}. 
For instance, if the representations are learned on sub-word levels, upon encountering sub-words from unseen relations, the parameters of the representation building network (encoder) can leverage sub-words shared between seen and unseen relations and thus, unseen relations will no longer have uninformed random vector representations.
%Several works on \dataset{WikiSQL} also encode column names on word level to yield vectors representing columns~\citep{incsql,stamp,syntaxsqlnet}.
% that is shared among both seen and unseen relations, the parameters of the representation building networks will be shared between seen and unseen relations and thus, unseen relations will no longer have uninformed random vector representations.
% Several works on question answering and semantic parsing (and other NLP tasks) have used such computed representations. 
% For example, several works on \dataset{WikiSQL} encode column names on word level to yield vectors representing columns~\citep{incsql,stamp,syntaxsqlnet}. 
% Some works on KGQA~\citep{yu2017improved,lukovnikov2017,maheshwari2018learning,golub2016} also decompose KG relations and/or entities to word and/or sub-word level.

Several works on question answering and semantic parsing (and other NLP tasks) have used such computed representations. For example,
Several works on \dataset{WikiSQL} also encode column names on word level to yield vectors representing columns~\citep{incsql,stamp,syntaxsqlnet}.
Some works on KGQA~\citep{yu2017improved,lukovnikov2017,maheshwari2018learning,golub2016} also decompose KG relations and/or entities to word and/or sub-word level.
In addition to sub-token level encoding, additional information about the tokens can be encoded and added to its representation.
For example, \citet{syntaxsqlnet}
also encodes table names and column data types together with column name words for their model for the \dataset{Spider} dataset.

\subsubsection{Translation based Parsing Algorithms %Semantic parsing using translation models.
}
\label{sec:translation:parsing}

In the case of purely translation-based semantic parsing, the \textit{parsing algorithm} centers around sequence decoding, which is usually done using greedy search or beam search as explained above.
%\dl{However, several }
%
%\paragraph{\af{Simple sequence-to-sequence models.}}
%A basic attention-based neural sequence-to-sequence model has been shown to perform well compared to previous approaches on some standard semantic parsing datasets~\citep{dong2016,jia2016}.
%\cite{golub2016} explore fully character-level encoding of NLQs, entities and \af{relations}, and use an attention-based \nc{encoder-}decoder architecture for \dataset{SimpleQuestions}.
% \dl{
% \paragraph{Entity Linking:}
% To avoid dealing with the large entity space and building representations for unseen entities, many approaches use external entity linking modules (see also Section~\ref{sec:pipeline}).
% With the entities identified, the main translation model can focus on the structure, predicates and operators in the logical form, filling in the already linked entities.
% %This allows the main semantic parsing model to focus mainly on predicatesbuilding the logical form and filling in the already linked entities.
% If the number of training examples is low, even for relatively small databases, entity linking may be required.
% }
\paragraph{Constrained Decoding}
A simple sequence-to-sequence model as described above does not exploit the formal nature of the target language. In fact, the output sequences to be generated follow strict grammatical rules that ensure the validity of the decoded expression, i.e.~that they can be executed over the given KG.
Thus, many translation based parsing algorithms exploit these grammatical rules by using grammar-based constraints during decoding in order to generate only valid logical forms. %(see Section~\ref{sec:constrained}). 
%\af{Such} grammatical rules behind logical forms dictate which sequences that can be produced by the decoder network are valid and can be \af{executed} over a \af{KG}.
%Most recent works on semantic parsing implement some form of constrained decoding that \af{inhibits} %can not
%\af{the generation} of invalid logical forms. \afrm{It can also be assumed that} 
%\af{[Note: Maybe remove the following sentence:]}
Using such constraints during training also help to focus learning and model capacity on the space of valid logical forms.
%\af{[TODO: The letter soumds a bit strange and needs further adaption.]}
%avoid wasting model capacity to learn the rules.
%
Depending on the specific application and dataset, reasonable assumptions concerning the logical form can be made. 
These choices are reflected in the logical form language definition.
As an example, %\af{that we have seen before}, 
in the case of \dataset{SimpleQuestions}, the assumption is that the logical from only consists of %only need to produce a 
single entity and a single relation. Therefore, if we were to solve \dataset{SimpleQuestions} using a translation model, we could constrain the decoder to choose only among all entities in the first time step and only among all relations in the second (instead of considering the full output vocabulary of all tokens possible in the formal query language) automatically terminate decoding thereafter.

For more general cases, we would like to express a wider range of formal query structures, but nevertheless apply some restrictions on the output tokens 
% \af{possible}
at a certain time step, depending on the sequence %query 
decoded so far.
For example, in the FunQL language definition used by \cite{cheng2017,cheng2019learning}, the $\mathsc{argmax}(x, r)$ function is restricted to have a relation symbol $r$ as second argument. Constrained decoding %decoding constraints 
can be trivially implemented by just considering % the most probable token that is in 
the set of allowed tokens as possible outputs at a certain time step.
Computing the allowed tokens for a certain time step however, can be more challenging, depending on the chosen logical form language.

\paragraph{Enhanced Decoding Procedures}
%\dltodo{moved from tree decoders!!!}
The two-stage Coarse2Fine decoder of \citet{dong2018}
can be seen as a middle-ground between a sequence decoder and a tree decoder. 
The decoder consists of two decoding stages: (1) decoding a query template and (2) filling in the specific details. 
%The proposed decoding scheme is similar to the other tree decoders: t
Compared to other tree decoders, the Coarse2Fine decoder also proceeds in a top-down breadth-first manner, but is restricted to have only two levels.
For cases when there is a limited number of query templates, \citet{dong2018} also investigate the use of a template classifier (instead of decoding the template) in the first decoding stage and evaluate on \dataset{WikiSQL}.
%This scheme can be advantageous w.r.t. tree decoders when there is a limited number of templates to choose from. In this case, the template decoding step can be replaced with a classifier, as \citep{dong2018} do for \dataset{WikiSQL}.
An additional improvement to the two-step decoding scheme is obtained by encoding the generated template using a bidirectional RNN, before using the output states to decode the details.
This allows to condition the generation of specific arguments of the template on the whole structure of the template. 
%In comparison, the normal depth-first sequence decoder can not have this ability, since it would require conditioning on the future. 
%Even though previously proposed tree decoders do not employ such intermediate encodings, it could also be incorporated in those using a breadth-first decoding order.

The work of \citet{cheng2018Weakly} trains and uses a translation model for semantic parsing. Using beam search, several logical forms are decoded from the translation model and additional ranking models (see Section~\ref{sec:weaksup}) are used to re-rank the logical forms in the beam. 
%proposes to use a basic ranking model as well as an inverse parsing model. The purpose of the inverse parsing model is to reconstruct the original question from the semantic parse. Together with the ranker and the inverse parser, the parsing algorithm of \citet{cheng2018Weakly} extends the simple decoding procedure commonly used to a procedure where beam search is used to generate a number of possible logical forms, where the beam is subsequently sorted using the scores from the basic ranker as well as the probabilities obtained from the inverse parser for the original input question.
%\dltodo{Add generative ranker here}

%\subsubsection{Model extensions}
%\label{sec:treedecoder}

\section{Emerging Trends}
\label{sec:trends}    
%The field of 
Question answering over knowledge graphs
%\af{KGQA}
has been an important area of research in the past decade. 
%Inviting, as well as
% Incoprorating and
% propagating advances from related fields like
% %IN the field of 
% deep learning, natural language understanding, transfer learning etc. KGQA has seen major improvements in the past couple of years, some of 
% the noteworthy of 
them being the use of query graph candidates~\citep{yih2015semantic, yu2017improved, maheshwari2018learning}, the use of neural symbolic machines~\citep{nsm}, the shift to answering multi-entity questions~\citep{luo2018knowledge}, the application of transfer learning~\citep{maheshwari2018learning}, and the proposal of differentiable query execution based weakly supervised models~\citep{yang2017differentiable}.
%Claims made \af{by} 
\citet{petrochuk2018simplequestions} suggest that the performance on \dataset{SimpleQuestions} %~cite{bordes2015large} 
is approaching an upper bound.
Further, as discussed in Section~\ref{sec:datasets}, there is a general shift of focus towards more complex logical forms, as is evident by recent datasets like ~\citep{trivedi2017lc, bao2016constraint,lcquad2}.
%\af{[Note: The letter seems repetitive to the point in the list above. Also we might not need to cite dataset papers again.]}\gm{an example or two to substantiate the claim might be a good idea.}
These advances lay a path for further improvements in the field, and the base for the %out
emerging trends and challenges we outline in the following.
%which we try summarising in this section.

\textbf{Query Complexity}
Comparative evaluations over \dataset{WebQuestions}, a dataset over \dataset{Freebase}, demonstrate the rise of performance of KGQA approaches over the years on the task~\citep{bao2016constraint}. 
Recently~\citet{petrochuk2018simplequestions} demonstrate ``\textit{that ambiguity in the data bounds} [the] \textit{performance at 83.4\% [over \dataset{SimpleQuestions}]}'' thereby suggesting that the progress over the task is further ahead than ordinarily perceived.
% For instance, the previously best performing baseline proposed by ~\citet{yu2017improved} is X\% closer to the upper limit, instead of 77\% as previously perceived.
For instance, the previously best performing baseline proposed by~\citet{yu2017improved} of 77\% accuracy over \dataset{SimpleQuestions}, can be perceived as 92.3\% (of 83.4) instead.
% is X\% closer to the upper limit, instead of 77\% as previously perceived.
Given that in \dataset{WebQuestions} where several KGQA systems have shown high performance, about 85\% of the quesitons are simple questions as well~\citep{bao2016constraint}, similar claims may be hold in this context, pending due investigations.

Knowledge graphs commonly used in KGQA community, as well as the formal query languages used to facilitate KGQA can support more nuanced information retrieval involving longer core chains, multiple triples, joins, unions, and filters etc.
These nuanced retrieval mechanisms are increasingly being regarded in the community as the next set of challenges. Recently released datasets such as \dataset{ComplexQuestions}~\citep{bao2016constraint}, \dataset{GraphQuestions}~\citep{su2016generating}, \dataset{LC-QuAD}~\citep{trivedi2017lc}, and \dataset{LC-QuAD 2}~\citep{lcquad2} explicitly focus on creating complex questions, with aggregates, constraints, and longer relation paths, over which the current systems do not perform very well, and there is a possibility of significant improvements. 

\textbf{Robustness}: In the past few years, deep learning based system have achieved state of the art performance over several tasks, however,
% In the past years, accelerated with the rise of generative adversarial networks~\citep{DBLP:conf/nips/GoodfellowPMXWOCB14}
% \af{[Note: GANs and adversarial examples are something different! We need to put the right citation here!]},
an increasing number of findings points out the brittleness of these systems.
For instance,~\citet{DBLP:conf/emnlp/JiaL17} demonstrate a drop of 35\%-75\% in F1 scores of sixteen models for the reading comprehension task trained over SQuAD~\citep{DBLP:conf/emnlp/RajpurkarZLL16}, by adversarially adding another sentence to the input paragraph (from which the system has to select the relevant span, given the question). 
Following, a new version of the aforementioned dataset was released comprising of \textit{unanswerable} questions~\citep{DBLP:conf/acl/RajpurkarJL18}, leading to more robust reading comprehension approaches~\citep{DBLP:journals/corr/abs-1808-05759, DBLP:conf/emnlp/KunduN18}. 
To the best of our knowledge, there hasn't been any work quantifying or improving the robustness of KGQA models. 
Such advances, would play an important role for the applicability
%applicable potential 
of KGQA systems in a production setting.
In our opinion, a good starting point for robustness inquiries would be to utilize recent general purpose adversarial input frameworks.
\citet{DBLP:conf/acl/SinghGR18}
propose a simple, generalisable way to generate semantically equivalent adversarial sentences. % Along similar lines, the paraphrase generation technique~\citep{} might be beneficial.

\textbf{Interoperability between KGs}: In discussing the numerous KGQA approaches in the previous section, we find that only a handful of techniques include datasets from both DBpedia and Freebase in their experiments, %}{Throughout the article, we treat QA datasets based on Freebase independent to the ones based on DBpedia} 
despite both of them being general purpose KGs. This is because the inherent data model underlying the two KGs differs a lot, %in the two datasets is much different,
which makes extracting \textit{(question, DBpedia answer)} pairs corrsponding to a set of \textit{(question, Freebase answer)} pairs %is 
(or vice versa) nontrivial, even after discounting the engineering efforts spent in migrating the query execution, and candidate generation sub-systems. 
%\af{[Note: is this really queation-answer pairs or NLQ-formal question pairs? Can we find a notation without parentheses?]}
This is best illustrated in~\citet{DBLP:conf/www/TanonVSSP16} where different hurdles and their solutions of migrating the knowledge from Freebase to Wikidata is discussed. Following a similar approach, ~\citet{DBLP:conf/semweb/DiefenbachTSM17} migrate the \dataset{SimpleQuestions} dataset to Wikidata, yielding 21,957 answerable questions over the KG.  \citet{DBLP:conf/coling/AzmySLI18} migrate it to DBpedia (October, 2016 release), and provide 43,086 answerable questions.
% This is best illustrated by~\citet{DBLP:conf/semweb/DiefenbachTSM17} who migrate SimpleQuestions %~\citep{bordes2015large} 
%  to Wikidata, and yield 21,957 answerable questions over the \af{[Note: freebase???]} KG.
%  \citet{DBLP:conf/coling/AzmySLI18} migrate it to DBpedia, and provide 43,086 answerable questions.

Correspondingly, interoperability of KGQA systems is another challenge, which only recently has drawn some attention in the community~\citep{Abbas2016WikiQAA}. 
The two fold challenge of (i) learning to identify multiple KG's artifacts %given a question
mentioned in a given NLQ, and (ii) learning multiple parse structures 
corresponding to the multiple KG's data models, while very difficult~\citep{DBLP:conf/ki/RinglerP17}, is partially helped by the latest (upcoming) DBpedia release\footnote{\url{http://downloads.dbpedia.org/repo/lts/wikidata/}}, whose data model is compatible with that of Wikidata's.
For a in-depth discussion on knowledge modeling strategies, and comparison of major largescale open KGs, we refer interested readers to~\citep{DBLP:journals/semweb/IsmayilovKALH18, DBLP:conf/ki/RinglerP17, DBLP:journals/semweb/FarberBMR18}.
% is just as difficult as learning to 
% This is best illustrated in~\citep{DBLP:conf/coling/AzmySLI18} who migrate \dataset{SimpleQuestions}~\citep{bordes2015large} to DBpedia

\textbf{Multilinguality}:
Multilinguality, i.e.~the ability to understand and answer questions in multiple languages
     % is integral to the idea of open knowledge,
is pivotal for a widespread acceptance and use of KGQA systems. 
With varying coverage, large parts of common knowledge graphs including DBpedia, Wikidata have multilingual surface forms corresponding to the resources, bridging a major challenge in enabling multilinguality in KGQA systems. 
% Thus multilinguality in KGQA is achievable slightly easily, when compared to other structured data sources. 
 
The QALD challenge, currently in its 9th iteration maintains multilingual question answering as one of its tasks. The dataset accompanying QALD-9\footnote{\url{https://project-hobbit.eu/challenges/qald-9-challenge/\#tasks}} multilingual QA over DBpedia task contains over 250 questions in upto eight languages including English, Spanish, German, Italian, French, Dutch, Romanian, Hindi and Farsi.
\citep{DBLP:journals/corr/abs-1803-00832} propose a non-neural \textit{generate and rank} approach with minimal language dependent components which can be replaced to support new languages. \citep{DBLP:conf/esws/RadoevZTG18} propose using a set of multilingual lexico-syntactic patterns to understand the intent of both French and English questions.
However, we still have to see a surge of multilinguality in data driven KGQA approaches. 
Since these approaches rely on supervised data to learn mappings between KG artifacts, and question tokens; the lack of large scale, multilingual, KGQA datasets inhibits these approaches. 

\section{Concluding Remarks}
\label{sec:conc}
Answering questions over knowledge graphs has emerged as an multi-disciplinary field of research, inviting insights and solutions from the semantic web, machine learning, and the natural language understanding community. In this article, we provide an overview of neural network based approaches for the task.

We broadly group existing approaches in three categories namely, (i) classification based approaches, where neural models are used to predict one of a fixed set of classes, given a question, (ii) ranking based approaches, where neural networks are used to compare different candidate logical forms with the question, to select the best-ranked one, and (iii) translation based where the network learns to translate natural language questions (NLQs) into their corresponding (executable) logical forms. Along with an overview of existing approaches, 
we also discuss some techniques used to weakly supervise the training of these models, which cope with the challenges risen due to a lack of logical forms in the training data.
We summarize existing datasets and tasks commonly used to benchmark these approaches, and note that the progress in the field has led to performance saturation over existing datasets such as \dataset{SimpleQuestions}, leading to an emergence of newer, more difficult challenges, as well as more powerful mechanisms to address these challenges.

Towards the end of the article, we discuss some of the emerging trends, and existing gaps in the KGQA research field, concluding that investigations and innovations in interoperability, multilinguality, and robustness of these approaches are required for impactful application of these systems.

\section{Acknowledgements}

We acknowledge support by the European Union H2020 grant Cleopatra (GA no.~812997).

\bibliography{references.bib}

\end{document}